\documentclass[
    a4paper,
    USenglish,
    authorcolumns,
    numberwithinsect,
    thm-restate,
]{lipics-v2021}

\usepackage[
    WOtheoremdefinitions,
]{auxlib}

\usepackage{algorithm}
\usepackage[noend]{algpseudocode}
\usepackage{nicefrac}

\usepackage[ampersand]{easylist}

\createcustomnote{todo}{backgroundcolor=purple}

\createcustomnote{peter}{backgroundcolor=green}
\createcustomnote{jonas}{backgroundcolor=yellow}
\createcustomnote{chr}{backgroundcolor=orange}

\ExplSyntaxOn
\ExplSyntaxOff

\def\lambdaContracting/{$\lambda$-contracting}
\def\lambdaPrimeContracting/{$\lambda'$-contracting}
\def\lambdaPrimePrimeContracting/{$\lambda''$-contracting}
\def\lambdaGathering/{$\lambda$-contracting gathering protocol}
\def\alphaBetaContracting/{$\left(\alpha,\beta\right)$-contracting}
\def\alphaBetaPrimeContracting/{$\left(\alpha',\beta'\right)$-contracting}
\def\alphaBetaPrimePrimeContracting/{$\left(\alpha'',\beta''\right)$-contracting}
\def\alphaBetaGathering/{$\left(\alpha,\beta\right)$-contracting gathering protocol}
\def\alphaBetaPrimeGathering/{$\left(\alpha',\beta'\right)$-contracting gathering protocol}
\def\lambdaPrimeGathering/{$\lambda'$-contracting gathering protocol}
\def\alphaCentered/{$\alpha$-centered}
\def\lambdaCentered/{$\lambda$-centered}
\def\clLambdaContracting/{collisionless $\lambda$-contracting}
\def\clLambdaPrimeContracting/{collisionless $\lambda'$-contracting}
\def\lambdaNearGathering/{$\lambda$-contracting near-gathering protocol}

\def\baseSegment/{\ensuremath{S_{\alpha}}}
\def\betaHalfSegment/{\ensuremath{S_{\alpha} \bigl(\frac{\beta}{2}\bigr)}}
\def\mainSegment/{\ensuremath{S_{\alpha} \bigl(\frac{\alpha \cdot \beta^2}{4}\bigr)}}
\def\intermediateSegment/{\ensuremath{S_{\alpha} \bigl(\frac{\alpha \cdot \beta}{4}\bigr)}}
\def\baseSegmentC/{\ensuremath{S_{\lambda \cdot \tau}}}

\def\baseSegmentLambda/{\ensuremath{S_{\lambda}}}
\def\betaHalfSegmentLambda/{\ensuremath{S_{\lambda} \bigl(\nicefrac{1}{2}\bigr)}}
\def\mainSegmentLambda/{\ensuremath{S_{\lambda} \bigl(\nicefrac{\lambda}{4}\bigr)}}
\def\intermediateSegmentLambda/{\ensuremath{S_{\lambda} \bigl(\nicefrac{\lambda}{4}\bigr)}}
\def\betaCSegmentLamda/{\ensuremath{S_{\alpha \cdot \tau} \bigl(\frac{\beta}{2}\bigr)}}
\def\baseSegmentCLambda/{\ensuremath{S_{\alpha \cdot \tau}}}

\def\baseCap/{\ensuremath{\mathrm{HSC}_{\alpha}}}
\def\betaHalfCap/{\ensuremath{\mathrm{HSC}_{\alpha} \bigl(\frac{\beta}{2}\bigr)}}
\def\mainCap/{\ensuremath{\mathrm{HSC}_{\alpha} \bigl(\frac{\alpha \cdot \beta^2}{4}\bigr)}}
\def\intermediateCap/{\ensuremath{\mathrm{HSC}_{\alpha} \bigl(\frac{\alpha \cdot \beta}{4}\bigr)}}

\def\baseCapLambda/{\ensuremath{\mathrm{HSC}_{\lambda}}}
\def\betaHalfCapLambda/{\ensuremath{\mathrm{HSC}_{\lambda} \bigl(\frac{1}{2}\bigr)}}
\def\mainCapLambda/{\ensuremath{\mathrm{HSC}_{\lambda} \bigl(\frac{\lambda}{4}\bigr)}}
\def\intermediateCapLambda/{\ensuremath{\mathrm{HSC}_{\lambda} \bigl(\frac{\lambda}{4}\bigr)}}

\def\cNG/{\ensuremath{c_{\mathrm{ng}}}}

\def\rNE/{\ensuremath{r^{NE}}}
\def\rBoundary/{\ensuremath{r^{Boundary}}}
\def\rCorner/{\ensuremath{r^{Corner}}}
\def\rTraverse/{\ensuremath{r^{Traverse}}}
\def\rGrid/{\ensuremath{r^{Grid}}}
\def\pOrth/{\ensuremath{p_{orth}}}

\def\gtc/{\textsc{Go-To-The-Center}}
\def\gtcShort/{\textsc{GtC}}
\def\gtmd/{\textsc{Go-To-The-Middle-Of-The-Diameter}}
\def\gtmdShort/{\textsc{GtMD}}
\def\gtcdmb/{\textsc{Go-To-The-Center-Of-The-Diameter-MinBox}}
\def\gtcdmbShort/{\textsc{GtCDMB}}

\def\LCM/{\textsc{LCM}}
\def\Look/{\textsc{Look}}
\def\Compute/{\textsc{Compute}}
\def\Move/{\textsc{Move}}

\def\gathering/{\textsc{Gathering}}
\def\nearGathering/{\textsc{Near-Gathering}}

\def\fsync/{$\mathcal{F}$\textsc{sync}}
\def\ssync/{$\mathcal{S}$\textsc{sync}}
\def\async/{$\mathcal{A}$\textsc{sync}}

\def\oblot/{$\mathcal{OBLOT}$}
\def\lumi/{$\mathcal{LUMI}$}

\newcommand*{\covered}{\operatorname{cov}}
\newcommand*{\hops}{\operatorname{hops}}
\newcommand*{\geomline}{\operatorname{line}}
\newcommand*{\mindist}{\operatorname{mindist}}
\newcommand*{\pos}{\operatorname{pos}}
\newcommand*{\sym}{\operatorname{sym}}
\newcommand*{\tail}{\operatorname{tail}}

\newcommand*{\iindex}{{(i)}}
\newcommand*{\einindex}{{(1)}}

\title{Forming Large Patterns with Local Robots in~the~OBLOT~Model}
\subtitle{arXiv-Version with extended Appendix}
\titlerunning{Forming Large Patterns with Local Robots in~the~OBLOT~Model (arXiv-Version)}

\author{Christopher Hahn}{Universität Hamburg, Germany}{christopher.hahn-1@uni-hamburg.de}{0009-0001-7617-6374}{}
\author{Jonas Harbig}{Paderborn University, Germany}{jonas.harbig@uni-paderborn.de}{0000-0003-3943-5979}{%
    This work was partially supported by the German Research Foundation(DFG) under the project number ME 872/14-1.
}
\author{Peter Kling}{Universität Hamburg, Germany}{peter.kling@uni-hamburg.de}{0000-0003-0000-8689}{}

\authorrunning{C. Hahn, J. Harbig, P. Kling}

\Copyright{%
    Christopher Hahn,
    Jonas Harbig and
    Peter Kling
}



\relatedversion{This is the version with an extended appendix\footnote{the following additions are made: \cref{sec:star-pattern} is slightly more detailed. \cref{app:auxiliary} includes the proof for \cref{lem:3connectedsectors}. \cref{sec:measuring-precision-appendix,sec:pseudocode,sec:example-execution} are added.} on arXiv. The related original version will be published in the conference proceeding of the 3rd Symposium on Algorithmic Foundations of Dynamic Networks (SAND 2024).}

\nolinenumbers
\hideLIPIcs

\def\refArxivSimulation{\cref{sec:example-execution}}
\def\refArxivPseudocode{\cref{sec:pseudocode}}

\def\refArxivMeasurement{\cref{sec:measuring-precision-appendix}}

\begin{document}

\maketitle

\begin{abstract}
{In the \emph{arbitrary pattern formation} problem, $n$ autonomous, mobile robots must form an arbitrary pattern $P \subseteq \R^2$.
The (deterministic) robots are typically assumed to be indistinguishable, disoriented, and unable to communicate.
An important distinction is whether robots have memory and/or a limited viewing range.
Previous work managed to form $P$ under a natural symmetry condition if robots have \emph{no memory but an unlimited viewing range}~\cite{DBLP:journals/tcs/YamashitaS10} or if robots have a \emph{limited viewing range but memory}~\cite{DBLP:conf/sirocco/YamauchiY13}.
In the latter case, $P$ is only formed in a shrunk version that has constant diameter.

Without memory and with limited viewing range, forming arbitrary patterns remains an open problem.
We provide a partial solution by showing that $P$ can be formed under the same symmetry condition if the robots' initial diameter is $\leq 1$.
Our protocol partitions $P$ into rotation-symmetric components and exploits the initial mutual visibility to form one cluster per component.
Using a careful placement of the clusters and their robots, we show that a cluster can move in a coordinated way through its component while \enquote{drawing} $P$ by dropping one robot per pattern coordinate.
}
\end{abstract}

{\section{Introduction}%
\label{sec:introduction}

Swarm robotics considers many, simple autonomous robots that must coordinate to reach a common goal.
Applications include exploration and rescue missions in hazardous environments (like the deep sea or space~\cite{kang2019marsbee}), medicine (for precise surgery or drug injection~\cite{DOI:10.1002/advs.202002203}), or biology (to model and understand the behavior of animal populations~\cite{DOI:10.1371/journal.pbio.3001269}).
While the degree of necessary cooperation varies between applications, a central aspect is almost always the deployment of robots to a given set of coordinates.

\paragraph{Model \& Problem}
The mentioned deployment aspect motivates the \emph{arbitrary pattern formation} problem, where a swarm of $n \in \N$ autonomous, mobile robots must form (in an arbitrary rotation and translation) a \emph{pattern} $P \subseteq \R^2$ of $\abs{P} = n$ \emph{coordinates}.
We assume the well-known \oblot/ ($\mathcal{OBL}$ivious rob$\mathcal{OT}$) model~\cite{DBLP:series/lncs/FlocchiniPS19} for (deterministic) point robots in $\R^2$ with the following characteristics:
Robots are \emph{oblivious} (have no memory), \emph{anonymous} (have no IDs), \emph{homogeneous} (execute the same protocol), and \emph{identical} (look the same).
They are also \emph{disoriented}, such that each robot perceives its surroundings in its own, local coordinate system that might be arbitrarily rotated and translated compared to other robots (and even vary over time).
A central feature of our work is that robots have a \emph{limited viewing range}: they perceive their surroundings up to a constant distance.
Without loss of generality, we normalize this viewing range to $1$.
We consider the \emph{fully-synchronous} (\fsync/) time model, where robots synchronously go through an \LCM/-cycle consisting of three phases: a \Look/- (observe surroundings), a \Compute/- (calculate target), and a \Move/- (move to target) phase.

A key aspect that determines whether a pattern can be formed is its symmetry.
For example, a swarm that starts as a perfectly regular $n$-gon cannot form an  arrow (which is, intuitively, less symmetric):
The robots may have identical local views and, thus, perform exactly the same computations and movements; the swarm would be forever trapped in a, possibly scaled, $n$-gon formation.
One can measure the symmetry of a pattern $P$ by its \emph{symmetricity} $\sym(P)$.
It counts how often $P$ covers itself when rotated full circle around its center (see~\cref{def:symmetricity}).
A swarm that starts with symmetricity $s$ can only form patterns whose symmetricity is a multiple of $s$~\cite{DBLP:journals/siamcomp/SuzukiY99, DBLP:journals/siamcomp/FujinagaYOKY15}.
This holds even for an \emph{unlimited} viewing range and for robots \emph{with memory}.
In fact, for oblivious robots (still with unlimited viewing range), these are \emph{exactly} the patterns that can be formed, even in an asynchronous setting~\cite{DBLP:journals/tcs/YamashitaS10, DBLP:journals/siamcomp/FujinagaYOKY15}.

Under limited viewing range, the situation is more elusive.
Robots \emph{with memory} can form a \emph{scaled} version of $P$ under the above symmetry condition~\cite{DBLP:conf/sirocco/YamauchiY13}.
Basically, the robots first form a \emph{near-gathering} (a formation in which robots have mutual visibility), use their memory to \enquote{maintain} the symmetricity, and then apply the protocol from~\cite{DBLP:journals/siamcomp/FujinagaYOKY15} to form a shrunk $P$ that fits into the viewing range.
In the \fsync/ setting, this holds even for \emph{non-rigid} moves (an adversary can stop robots during their move).
On the negative side, oblivious robots with \emph{non-rigid} moves cannot always form $P$, even if the symmetry condition holds~\cite{DBLP:conf/sirocco/YamauchiY13}.

It remains open whether the patterns that can be formed by oblivious robots with limited viewing range (and rigid movements) are also characterized by the symmetry condition.

\paragraph{Our Contribution}
We make a decisive step towards characterizing patterns that \emph{oblivious} robots with \emph{limited viewing range} can form without \emph{down-scaling}.
Our main result is the following \lcnamecref{thm:main-result} (see \cref{sec:model} for formal definitions):
\begin{theorem}[{name=, restate=[name=restated]thmMainResult}]%
\label{thm:main-result}
A connected pattern $P$ can be formed by $\abs{P}$ oblivious \oblot/ robots with limited viewing range in the \fsync/ model from a near-gathering $I$ if and only if $\sym(I) \mid \sym(P)$.
The formation takes $\ldauOmicron{n}$ rounds, which is worst-case optimal.
\end{theorem}

Starting from a near-gathering avoids another challenging open problem:
Can we reach a near gathering from any connected formation \emph{without increasing the symmetricity}.
If that were the case, together with \cref{thm:main-result} it would show that (under rigid movements) obliviousness and limited viewing range do not weaken the robots' pattern formation abilities.
Note that a recent near-gathering protocol for our model~\cite{DBLP:conf/opodis/CastenowH0KKH22} avoids \emph{collisions} (two robots moving to the same spot), a major cause of symmetricity increase for most gathering protocols.

Requiring that $P$ is connected (see \cref{sec:model}) is natural for robots of limited viewing range.
However, our technique can be adapted to form disconnected patterns, as long as they contain a connected component of size $\geq 3$  (see \cref{sec:conclusion} for a brief discussion).

In a nutshell, our protocol partitions the input pattern $P$ into $\sym(P)$ rotation-symmetric \emph{components}.
Using the initial mutual visibility, we let the robots form one cluster, called \emph{drawing formation}, per component.
Such a drawing formation relies on a careful placement of its contained robots to store information about the component it is responsible for and to coordinate its movement.
We show how the drawing formation's robots can compute and coordinately move along a deliberately constructed path through the component in order to \enquote{draw} the pattern by dropping one robot at each contained coordinate.

\paragraph{Further Related Work}
The arbitrary pattern formation problem has been considered in numerous settings and variants.
To name just a few, there are results for pattern formation
\begin{itemize*}[afterlabel=, label=]
\item on a grid~\cite{DBLP:journals/tcs/BoseAKS20},
\item with obstructed view~\cite{DBLP:journals/tcs/BoseKAS21},
\item with axis agreement~\cite{DBLP:journals/tcs/FlocchiniPSW08},
\item for robots without a common chirality~\cite{DBLP:journals/dc/CiceroneSN19},
\item for pattern sequences~\cite{DBLP:journals/dc/0001FSY15}, and
\item in three dimensional space~\cite{DBLP:conf/podc/YamauchiUY16}.
\end{itemize*}
See~\cite{DOI:10.1088/1742-6596/473/1/012016} for a survey on pattern formation.
A more recent and general overview of results and open problems in swarm robotics and related areas can be found in~\cite{DBLP:series/lncs/11340}.

There is also work dedicated to forming a specific pattern like
\begin{itemize*}[afterlabel=, label=]
\item a point~\cite{DBLP:conf/spaa/DegenerKLHPW11, DBLP:conf/opodis/CastenowH0KKH22, DBLP:journals/tcs/CastenowFHJH20} (gathering),
\item an arbitrarily tight near-gathering~\cite{DBLP:journals/siamcomp/CohenP05, DBLP:conf/podc/KirkpatrickKNPS21} (convergence), and
\item a uniform circle~\cite{DBLP:series/lncs/Viglietta19, DBLP:conf/icdcit/MondalC20, DBLP:journals/dc/FlocchiniPSV17}.
\end{itemize*}
Again, a rather up-to-date and good overview can be found in~\cite{DBLP:series/lncs/11340}.

Somewhat different in spirit but in our context relevant is~\cite{DBLP:journals/dc/LunaFSV22}.
The authors show how three or more robots with limited viewing range but \emph{arbitrarily precise} sensors can form a \emph{TuringMobile} to simulate a Turing machine that can, e.g., store and process real numbers.
To showcase the model's power, the authors provide, amongst others, a pattern formation protocol for any dimension and an initially disconnected swarm, but under the strong requirement that (some) robots form initially a TuringMobile.
Note that while we use robot placement to encode information, we do this in an inherently discrete way, requiring a sensor precision of order only $\min\set{1/\sqrt{\abs{P}}, \mindist(P), 1/\sym(P)}$.
A precision of order $1/\sqrt{\abs{P}}$ is already required to measure distances in any near-gathering of $\abs{P}$ robots.
Similarly, to form $P$, robots must naturally be able to measure the minimal distance $\mindist(P)$ that occurs in $P$.
The final term stems from the fact that our drawing starts from a near-gathering of symmetricity $\sym(P)$, in which robot distances are of order $\LdauOmicron{1 / \sym(P)}$.
In \refArxivMeasurement, we add a discussion of the required measuring precision.

\paragraph{Outline}
\Cref{sec:model} introduces preliminaries like further notions and notation.
\Cref{sec:general_approach} contains the major part of our protocol description and its analysis.
That \lcnamecref{sec:general_approach} formalizes notions like drawing formations or drawing paths and details how we coordinate the robots that form a drawing formation.
At the \lcnamecref{sec:general_approach}'s end, we prove \cref{thm:main-result} under the assumptions that there is a drawing path that adheres to certain conditions (namely \cref{def:compatibiliy}) and that $\sym(P) < \abs{P}/2$.
The construction of such a drawing path is subject of \cref{sec:dp-implementation}.
We conclude with a small discussion and open problems in \cref{sec:conclusion}.
}
{\section{Preliminaries \& Notation}%
\label{sec:model}

This \lcnamecref{sec:model} extends the model and problem description from \cref{sec:introduction}.

\paragraph{Geometric Notation}
For two points $p, q \in \R^2$ we define $\dist(p, q) = \norm{p - q}_2$ as their Euclidean distance.
We extend this notation in the natural way to sets $S \subseteq \R^2$, such that, e.g., $\dist(p, S) = \min\set{dist(p, q) | q \in S}$.
We use a set-like notation for sequences $S = \intoo{s_i}_{i=1}^{n}$, like $p_1 \in S$, $S \subseteq \R^2$, or $\dist(p, S)$.
The minimal distance between two points in a set (or sequence) $S \subseteq \R^2$ is $\mindist(S) \coloneqq \min\set{\dist(p, q) | p, q \in S, p \neq q}$.
For $p \in \R^2$ and $r \in \R$ the set $\cB(p, r) = \set{q \in \R^2 | \dist(p, q) < r}$ denotes the open ball around $p$ with radius $r$.

For a set $S \subseteq \R^2$ we write
\begin{itemize*}[afterlabel=, label=]
\item its power set as $\cP(S)$,
\item its closure as $\overline{S}$, and
\item its boundary as
    \begin{math}
    \partial S
    =
    \overline{S} \cap \overline{\R^2 \setminus S}
    \end{math}.
\end{itemize*}
To highlight the usage of directions (in contrast to points), we use vector notation like $\vec{u}, \vec{v} \in \R^2$.
Let $G_S = (S, E)$ with $E = \set{ \set{p, q} \subseteq S | \dist(p, q) \in \intoc{0, 1} }$ be the \emph{unit disc graph} of $S$.
Then $S$ is \emph{connected} if $G_S$ is connected and $p, q \in \R^2$ are \emph{connected} if $\set{p, q}$ is connected.
We use $\angle(\vec{u}, \vec{v}) \in \intoc{-\pi, \pi}$ for the signed angle between $\vec{u}$ and $\vec{v}$.
If not stated otherwise, explicit coordinates for points $p = (r, \phi) \in \R^2$ are given in \emph{polar coordinates}.

\paragraph{Patterns \& Configurations}
Remember that we consider the \emph{arbitrary pattern formation problem}: a swarm $\cR$ of $n \coloneqq \abs{\cR} \in \N$ \oblot/ robots with a viewing range of $1$ must form a target pattern $P \subseteq \R^2$ of $\abs{P} = n$ coordinates.
Since robots are oblivious, we use the standard assumption that, each round, they receive $P$ as their sole input in an arbitrary but fixed coordinate system (i.e., robots receive the exact same numerical values).

Since robots are deterministic and indistinguishable, the \emph{configuration} at any time is uniquely described by the robots' positions $\pos(\cR) = \set{ \pos(r) | r \in \cR } \subseteq \R^2$.
If the robot identity is irrelevant for the matter at hand, we identify $r \in \cR$ with the position $\pos(r)$ and $\cR$ with the configuration $\pos(\cR)$.
A \emph{near-gathering} is a configuration of diameter $\leq 1$.

W.l.o.g., we assume that $P$'s smallest enclosing circle is centered at the origin (otherwise, robots translate $P$ accordingly).
We measure $P$'s symmetry via its \emph{symmetricity}:
\begin{definition}[Symmetricity~\cite{DBLP:journals/siamcomp/FujinagaYOKY15}]%
\label{def:symmetricity}
Consider a set $P \subseteq \R^2$ whose smallest enclosing circle is centered at $c \in \R^2$.
A \emph{$m$-regular} partition of $P$ is a partition of $P$ into $k = \abs{P}/m$ regular $m$-gons with common center $c$.
The \emph{symmetricity} of $P$ is defined as
\begin{math}
\sym(P)
\coloneqq
\max\set{ m \in \N | \text{there is a $m$-regular partition of $P$} }
\end{math}.
\end{definition}
In \cref{def:symmetricity}, a single point is considered a $1$-gon with an arbitrary center.
Thus, any $P$ has a $1$-regular partition.
Note that, if the origin is an element of $P$, then $\sym(P) = 1$.\footnote{
    One might assume a $n$-gon together with its center forms a rather symmetric set of size $n+1$.
    But robots can easily break the perceived symmetry, since the center robot basically functions as a leader.
}
At some places, we use the shorthand $s_P$ for the symmetricity $\sym(P)$ of a set $P$

Symmetricity allows us to characterize patterns that can be formed by synchronous, oblivious robots with an unlimited viewing range:
\begin{theorem}[{Symmetry Condition, \cite[Theorem~1]{DBLP:journals/siamcomp/FujinagaYOKY15}}]%
\label{thm:symmetry-condition}
A pattern $P$ can be formed by $\abs{P}$ oblivious \oblot/ robots with unlimited viewing range in the \fsync/ model from configuration $I$ if and only if $\sym(I) \mid \sym(P)$.
\end{theorem}

\paragraph{Further Time Models}
Remember that we assume the \emph{fully-synchronous} time model (\fsync/), where each \emph{round} robots synchronously execute the \Look/-, \Compute/-, and \Move/-phases of their \LCM/-cycle.
Two other natural models are the \emph{semi-synchronous} time model (\ssync/; a subset of robots is active each round and executes its phases synchronously) and the \emph{asynchronous} time model (\async/; robots execute phases completely asynchronously).
}
{\section{Forming Patterns via Drawing}%
\label{sec:general_approach}

Given a pattern $P$ of symmetricity $s_P \in \N$, we \enquote{draw} $P$ using $s_P$ \emph{drawing formations}.
Each drawing formation consists of a carefully arranged subset of \emph{state robots} and is responsible to form one of $s_P$ symmetric subpatterns $P' \subseteq P$.
The state robots' careful placement enables them to coordinately move through $P'$ along a specific \emph{drawing path}.
While doing so, some state robots are \emph{dropped} at nearby pattern coordinates to form $P'$.

We start in \cref{sec:drawing-formations} by formalizing the idea of drawing formations and related concepts.
\Cref{sec:df-implementation} details the legal arrangements of state robots in a drawing formation and shows how we can use this to coordinate state robots.
Equipped with this coordination, \cref{sec:drawing-symmetric-component} formalizes the drawing path $\bm{v}$ and what properties a drawing formation $F$ must have in order to traverse $\bm{v}$ while suitably dropping robots.
Afterward, \cref{sec:df-pattern} shows how we can create $s_P$ different drawing formations and use them to draw suitable, symmetric subpatterns of $P$.
Finally, \cref{sec:main-result-proof} puts everything together to prove \cref{thm:main-result}.
We often define algorithms implicitly during the proofs.
To better illustrate the algorithms, we give high-level pseudocode in \refArxivPseudocode.

\subsection{Drawing Formations \& Movement}%
\label{sec:drawing-formations}

We first define the notion \emph{drawing hull}, representing the general shape of a drawing formation.
\begin{definition}[Drawing Hull]%
\label{def:drawing-hull}
A \emph{drawing hull} $H = (a, \vec{d}, \phi, \Delta)$ consists of an \emph{anchor} $a \in \R^2$, a \emph{direction} $\vec{d} \in \R^2$ with $\norm{\vec{d}}_2 = 1$, a \emph{span} $\phi \in \intoc{0, \pi/3}$, and a \emph{diameter} $\Delta \in \intoc{0, 1}$.
\end{definition}
As illustrated in \cref{fig:drawing-hull}, one should think of a drawing hull $H = (a, \vec{d}, \phi, \Delta)$ as the point set $\set{x \in \R^2 | \dist(x, a) \leq \Delta \land \angle(\vec{d}, x - a) \in \intco{0, \phi}}$.\footnote{
    Note that $\phi \leq \pi/3$ ensures that $\Delta$ is indeed the diameter of the point set $H$. 
}
With this in mind, we sometimes abuse notation and identify $H$ with this set to write, e.g., $\pos(r) \in H$ for a robot $r$.

A \emph{drawing formation} is defined by a drawing hull and all robots contained in it.
These robots form a tight cluster whose exact placement inside the hull (the drawing formation's \emph{state}) allows us to coordinate their movement (see \cref{sec:df-implementation}).
\begin{definition}[Drawing Formation]%
\label{def:drawing-formation}
A \emph{drawing formation} $F = (H_F, \cR_F)$ consists of a drawing hull $H_F$ and the robot set $\cR_F \coloneqq \set{ r \in \cR | \pos(r) \in H_F }$.
We call $r \in \cR_F$ a \emph{state robot} of $F$ and $\cS_F \coloneqq \pos(\cR_F)$ the \emph{state} of $F$.
The \emph{size} of $F$ is $\abs{\cR_F}$.
\end{definition}

We sometimes identify a drawing formation with its hull, allowing us to, e.g., speak of a drawing formation's anchor or diameter.

A drawing formation $F$ forms a given pattern by \enquote{moving} $F$ along a specific \emph{drawing path} (see \cref{sec:drawing-symmetric-component}) that visits all pattern coordinates, dropping one state robot per pattern coordinate along the way.
The following \lcnamecref{def:df-move} formalizes such \emph{moves} (see \cref{fig:df-movement} for an illustration).
\begin{definition}[Move]%
\label{def:df-move}
Consider a drawing formation $F= (H_F, \cR_F)$ with drawing hull $H_F = (p, \vec{d}, \phi)$ in configuration $\cR$.
Let $\cR'$ denote the configuration after the next \LCM/ cycle.
We say $F$ \emph{moves} from $p$ (in configuration $\cR$) to $p'$ (in configuration $\cR'$) if a state robot subset $\cR_{F'} \subseteq \cR_F$ of $F$ forms a drawing formation $F' = \intoo[\big]{(p', \vec{d}, \phi), \cR_{F'}}$ in configuration $\cR'$.
We call the robots $\cR_F \setminus \cR_{F'}$ \emph{dropped} robots.
\end{definition}

When moving from one drawing path vertex to the next, the remaining state robots change state (their placement in the drawing formation) to encode the progress on the drawing path.
To ensure that a drawing formation can adopt any (reasonable) state after a movement, we restrict its movement distance to $1 - \Delta$ (s.t.~each state robot can reach any other location in the resulting drawing formation of diameter $\Delta$).
\begin{observation}%
\label{obs:moving}
Consider a drawing formation $F$ of diameter $\Delta$ that moves from position $p$ to $p'$.
If $\dist(p, p') \leq 1 - \Delta$, the robots that are not dropped can form any state in the resulting drawing formation.
\end{observation}

\subsection{States of a Drawing Formation}%
\label{sec:df-implementation}

Given a target pattern $P$, our protocol considers only drawing formations $F$ with fixed span $\phi = 2\pi / \sym(P)$ (depending only on $P$) and fixed diameter $\Delta$ (constant).
Moving $F$ between vertices of the drawing path (see \cref{sec:drawing-symmetric-component}) requires a coordinated movement of $F$'s state robots.
To achieve this, any robot must
\begin{enumerate}
\item\label{enm:req:deducehull} decide whether it is one of $F$'s state robots and, if so,
\item\label{enm:req:deduceprogress} know the current progress on the drawing path.
\end{enumerate}
To achieve~(\ref{enm:req:deducehull}), we use a careful placement of three \emph{defining robots} that allows any robot $r$ that sees them to deduce the remaining hull parameters (anchor and direction); once all four hull parameters are known, $r$ can compute the hull $H_F$ and decide whether it lies inside $H_F$ or not.
To achieve~(\ref{enm:req:deduceprogress}), we require that any additional state robots are placed on an $\epsilon$-grid ($\epsilon > 0$ fixed, depending only on $P$) that is aligned with the defining robots; using an arbitrary but fixed enumeration scheme for $\ell$ robots on such a grid, all state robots can agree on the same ordering of states and use it (in combination with $F$'s size $\ell$) to encode the progress on the drawing path.

\paragraph{Legal States}
We continue to formalize this idea for a given parameter $\epsilon > 0$.
The placement of the defining robots $r_1$, $r_2$, and $r_3$ of a drawing formation $F$ with anchor $a$ and direction $\vec{d}$ is as follows:
\begin{enumerate}
\item $r_1$ is at the anchor $a$,
\item $r_2$ is at distance $\epsilon$ in direction $\vec{d}$ from anchor $a$, and
\item $r_3$ is at distance $\in \set{2\epsilon, 4\epsilon, \dots}$ in direction $\vec{d}$ from $r_2$.
\end{enumerate}
Further state robots (if any) must be placed on the non-negative $2\epsilon$-grid with origin $r_2$ and whose $x$-axis is aligned with $\vec{d}$ (see \cref{fig:epsilon-granular-drawing-formation}).

The robot pair $\set{r_1, r_2}$ can be identified since they are the only state robots with distance $\epsilon$.
And since $r_3$ at distance $\geq 2\epsilon$ is closer to $r_2$ than to $r_1$, robots can distinguish $r_1$ from $r_2$, from which they can infer both the hull's anchor and direction.

We get the following set of potential state robot locations:
\begin{definition}[$\epsilon$-Granular Locations]%
\label{def:epsgranular-locations}
Consider a drawing formation $F = (H_F, \cR_F)$ with anchor $a$ and direction $\vec{d}$.
Let $\vec{d}_{\bot}$ be the unit vector with $\angle(\vec{d}, \vec{d}_{\bot}) = \nicefrac{\pi}{2}$.
The set of \emph{$\epsilon$-granular locations} of $F$ is
\begin{dmath}
L_F(\epsilon)
\coloneqq
\set{ a, a + (1 + 2i) \epsilon \cdot \vec{d} + 2j\epsilon \cdot \vec{d}_{\bot} | i, j \in \N_0 } \cap H_F
\end{dmath}.
\end{definition}

States (i.e., state robot placements) considered legal by our protocol consist of all possible placements on $\epsilon$-granular locations with the mentioned restrictions on the three defining robots' positions.
\begin{definition}[$\epsilon$-Granular States]%
\label{def:epsilon-granular-state}
Consider a drawing formation $F$ with anchor $a$ and direction $\vec{d}$.
The set of \emph{$\epsilon$-granular states} of $F$ is 
\begin{dmath*}
A_F(\epsilon)
\coloneqq
\Set{\cS \cup  \cT | \cS \in \cP(L_F), \cT \in T(\epsilon)}
\end{dmath*}.
with
\begin{math}
T(\epsilon)
\coloneqq
\bigcup_{i=1}^{\floor{(\nicefrac{\Delta}{\epsilon}-1)/2}} \Set{\{a, a + \epsilon \cdot \vec{d}, a + (1 + 2i) \epsilon \cdot \vec{d}\}}
\end{math} being the sets of defining robots.
For $\ell \in \N$ we define
\begin{math}
A_F^{\ell}(\epsilon)
\coloneqq
\set{ \cS \in A_F(\epsilon) | \abs{\cS} = \ell}
\end{math}
as the set of all $\epsilon$-granular states of $F$ that can be adopted with $\ell$ state robots.
\end{definition}

Our protocol considers only drawing formations that adhere to the above restrictions, leading us to the following \lcnamecref{def:epsilon-granular-df}:
\begin{definition}[$\epsilon$-Granular Drawing Formation]%
\label{def:epsilon-granular-df}
A drawing formation $F$ is \emph{$\epsilon$-granular} if $F$ is in an $\epsilon$-granular state and if $F$'s state robots know\footnote{
    The parameters are either hard-coded into the protocol or can be computed from the target pattern $P$.
} the fixed parameters $\epsilon$, $\Delta$, and $\phi$.
\end{definition}

Note that all subsets of robots in the current configuration that fulfill the definition above are $\epsilon$-granular drawing formations.
We require that $r \in F$ knows the parameter $\epsilon, \Delta$ and $\phi$ of $F$.
Therefore $r \in F$ can check all subsets in its viewing range whether they are drawing formations with these parameters.
With a viewing range of $1 \geq \Delta$, $r$ observes all robots in $F$ and can compute the drawing hull of $F$.

\begin{observation}%
\label{lem:valid-df}
Let $F$ be a $\epsilon$-granular drawing formation.
All state robots in $F$ can compute the anchor of $F$ and $\vec{d}$.
\end{observation}

\begin{figure}
    \begin{subfigure}[t]{0.28\linewidth}
        \centering\includegraphics[page=7,width=\textwidth]{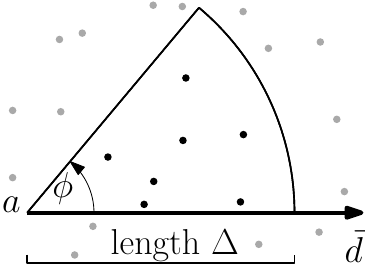}
        \caption{
            Drawing formation $F$ with drawing hull $H_F$ and state robots $r \in \cR_F$ (black) and other robots $r \notin \cR_F$ (gray).
        }\label{fig:drawing-hull}
        \end{subfigure}
        \hfill
    \begin{subfigure}[t]{0.33\linewidth}
        \centering\includegraphics[page=8,width=0.65\textwidth]{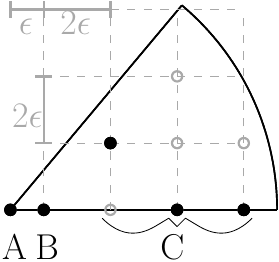}
        \caption{
            $\epsilon$-granular drawing formation; locations A and B must each contain a robot, the locations at C must in sum contain $\geq 1$ robots.
        }\label{fig:epsilon-granular-drawing-formation}
        \end{subfigure}
        \hfill
    \begin{subfigure}[t]{0.32\linewidth}
            \centering\includegraphics[page=9,width=\textwidth]{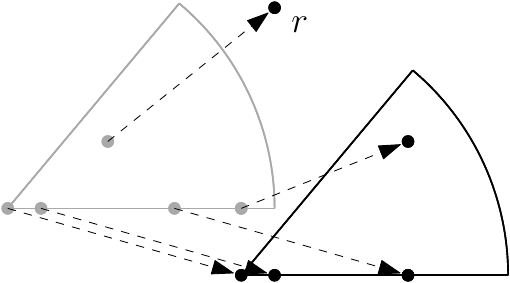}
            \caption{
                Movement of a drawing formation where robot $r$ is dropped.
            }\label{fig:df-movement}
            \end{subfigure}
            \hfill
    \caption{}
\end{figure}

As a final property, we only want non-overlapping drawing formations.
If two drawing hulls were overlapping, a robot might be state robot in two drawing formations that move in different directions.
\begin{definition}[Validity]%
\label{def:valid-df}
An $\epsilon$-granular drawing formation $F = (H_F, \cR_F)$ is \emph{valid in configuration $\cR \supseteq \cR_F$} if for any other $\epsilon$-granular drawing formation $F' = (H_{F'}, \cR_{F'})$ we have $H_F \cap H_{F'} = \emptyset$.
\end{definition}

\paragraph{Counting via States}
Given an $\epsilon$-granular drawing formation $F$ of size $\ell$ (i.e., consisting of $\ell$ state robots), we can easily enumerate $A_F^{\ell}(\epsilon)$ in a way that depends solely on the relative positions of its locations.
In particular, all state robots can use this enumeration and thus agree on the order of states, which basically equips the state robots with a shared counter.
A concrete implementation is depicted in \refArxivSimulation.
\begin{definition}[$i$-th State]%
\label{def:df-state-enum}
Using an arbitrary, unique state enumeration on $A_F^{\ell}(\epsilon)$ that depends solely on the relative position of locations, for $i \in \set{1, 2, \dots, \abs{A_F^{\ell}(\epsilon)}}$ we define the \emph{$i$-th state} of $F$ as the corresponding state in this enumeration.
\end{definition}

We conclude with a lower bound on the number of states that an $\epsilon$-granular drawing formation may have.
\begin{lemma}%
\label{lem:df-state-size}
Consider a drawing formation $F$ with hull diameter $\Delta$ and span $\phi \in \intoc{0, \pi/3}$.
We have
\begin{math}
\abs{A_F^3(\epsilon)}
=
\floor{(\Delta / \epsilon - 1) / 2}
\end{math}
and
\begin{math}
\abs{A_F^{\ell}(\epsilon)}
=
\ldauOmega{\Delta^3 \cdot \phi / \epsilon^3}
\end{math}
for $4 \leq \ell \leq |L_F(\epsilon)| - 1$.
\end{lemma}
\begin{proof}
For $\ell = 3$ robots, only the third defining robot $r_3$ can choose between multiple locations; the first and second defining robots $r_1$ and $r_2$ have a fixed location inside $H_F$.
Since $r_3$ can be placed on any of the locations of the form $a + (1 + 2i)\epsilon \cdot \vec{d}$ for $i \in \N$ that lies in $H_F$, we get $\abs{A_F^3} = \floor{(\Delta / \epsilon - 1) / 2}$.

For $\ell = 4$, observe that there are $k \coloneqq \abs{L_F(\epsilon)} = \ldauOmega{\Delta^2\cdot \phi / \epsilon^2}$ $\epsilon$-granular locations in $F$, since the $2\epsilon$-grid allows for $\ldauOmega{1/\epsilon^2}$ many locations per unit area and the total area covered by $F$'s hull is
\begin{math}
\pi \cdot \Delta^2 \cdot \phi / (2\pi)
=
\Delta^2 \cdot \phi / 2
\end{math}.
Again, the first two defining robots have a fixed location, while the third defining robot may occupy one of $\ldauOmega{\Delta / \epsilon}$ many locations.
The remaining $\ell - 3 \geq 1$ robots can be arranged on the remaining $k - 3$ locations in $\binom{k-3}{\ell-3}$ ways.
By the \lcnamecref{lem:df-state-size}'s restriction on $\ell$ we have $\ell - 3 \geq 4$ and $\ell - 3 \leq k - 4$, such that $\binom{k-3}{\ell-3} \geq \binom{k-3}{k-4} = \binom{k-3}{1} = \ldauOmega{\Delta^2 \cdot \phi / \epsilon^2}$.
Together, we get the desired bound.
\end{proof}

\subsection{Drawing a Pattern via a Drawing Path}%
\label{sec:drawing-symmetric-component}

This \lcnamecref{sec:drawing-symmetric-component} introduces the \emph{drawing path} of a (sub-) pattern $P$, which is a path in $\R^2$ that visits all pattern coordinates.
This path should allow a drawing formation to move along its vertices while dropping one state robot per pattern coordinate along the way to form $P$.
An instructive illustration of the idea can be found in \refArxivSimulation.

A drawing path $\bm{v}$ has a parameter $\delta$ that controls the maximal distance between consecutive vertices as well as between each pattern coordinate and the path.
Moreover, $\bm{v}$ must depend only on $P$, such that oblivious robots can all recalculate $\bm{v}$ each \LCM/-cycle.
\begin{definition}[Drawing Path]%
\label{def:drawing-path}
Consider any pattern $P$.
A path $\bm{v} = \intoo{v_j}_{j=1}^{k}$ of $k$ vertices $v_j \in \R^2$ is a \emph{$\delta$-drawing path} of $P$ if
\begin{enumerate}
\item $\bm{v}$ can be calculated from $P$,
\item $\forall p \in P\colon \dist(p, \bm{v}) \leq 1 - \delta$, and
\item $\forall j \in \set{1, 2, \dots, k-1}\colon \dist(v_j, v_{j+1}) \leq 1 - \delta$.
\end{enumerate}
\end{definition}
Choosing $\delta$ equal to the diameter of a drawing formation $F$ enables $F$ to traverse $\bm{v}$ while forming any state (\cref{obs:moving}).
We omit $\delta$ if it is irrelevant for the matter at hand.

When traversing $\bm{v}$, we want a drawing formation to drop a robot at pattern coordinate $p \in P$ when leaving the latest vertex $v_j$ that is close enough to $p$.
This ensures that, at any time after being dropped, the dropped robot has a distance of at least $1 - \delta$ to the drawing formation that dropped it.
We say $p$ is \emph{covered} by vertex $v_j$.
\begin{definition}[Covered Coordinates]%
\label{def:covered-coordinate}
Consider a $\delta$-drawing path $\bm{v} = \intoo{v_j}_{j=1}^{k}$ of a pattern $P$.
Coordinate $p \in P$ is \emph{covered} by vertex $v_j$ if $j \in \set{1, 2, \dots, k}$ is the maximal index for which $\dist(p, v_j) \leq 1 - \delta$.
Let $\covered(v_j)$ denote the set of all coordinates covered by $v_j$.
\end{definition}
We extend $\covered(\bullet)$ in the natural way to subsequence $\bm{v'}$ of $\bm{v}$, such that $\covered(\bm{v'}) = \bigcup_{u \in \bm{v'}} \covered(u)$.

Care must be taken once a drawing formation dropped so many robots that it reached size $\ell = 3$:
It must not drop further robots before the path's end, since the remaining two robots would no longer form a drawing formation and could not coordinate (see \cref{sec:df-implementation}).
We capture this (possibly non-existent) path region in the following \cref{def:dp-tail}.
\begin{definition}[Tail of a Drawing Path]%
\label{def:dp-tail}
The \emph{tail} $\tail(\bm{v})$ of a drawing path $\bm{v} = \intoo{v_j}_{j=1}^{k}$ is the longest suffix $\intoo{v_j}_{j=s}^{k}$ s.t.~$\sum_{j=s}^{k} \abs{\covered(v_j)} < 4$.
\end{definition}

As a final notion, we declare when a drawing formation and a drawing path are \emph{compatible} (i.e., can be used to form the path's pattern).
Here, we use $\hops(v_s, v_t)$ to denote the number of edges between two path vertices $v_s, v_t \in \bm{v}$.
\begin{definition}[Compatibility]%
\label{def:compatibiliy}
An $\epsilon$-granular drawing formation $F$ with diameter $\Delta$ and span $\Phi$ is \emph{compatible} with a $\delta$-drawing path $\bm{v} = \intoo{v_j}_{j=1}^{k}$ of a pattern $P$ if
\begin{enumerate}
\item\label[property]{enm:compreq:params} $\epsilon < \mindist(P)$ and $\Delta \leq \delta$,
\item\label[property]{enm:compreq:bodytraversable} $\forall s < t\text{ s.t. }\bigcup_{j=s}^{t-1} \covered(v_j) = \emptyset\colon \hops(v_s, v_t) \leq \abs{A_F^{4}(\epsilon)}$,
\item\label[property]{enm:compreq:tailtraversable} $\abs{\tail(\bm{v})} \leq \abs{A_F^{3}(\epsilon)}$, and
\item\label[property]{enm:compreq:finalstep} $\abs{\covered(\tail(v))} = 3$ and $\covered(\tail(\bm{v})) \subseteq \cB(v_k, 1)$ .
\end{enumerate}
\end{definition}
\Cref{enm:compreq:params} ensures that the distance $\epsilon$ (identifying $F$'s defining robots) does not occur in $P$ and that $F$ can traverse $\bm{v}$ (by \cref{obs:moving}).
\Cref{enm:compreq:bodytraversable} requires that, after dropping a robot, the state space $A_F^{\ell} \supseteq A_F^{4}$ of the remaining $\ell$ robots is large enough to encode the progress towards the next vertex where a robot is dropped.
These two properties are used in \cref{lem:drawing-body} to prove that $F$ can traverse the non-tail of $\bm{v}$ while appropriately dropping robots.

\Cref{lem:drawing-tail} uses \cref{enm:compreq:tailtraversable} to traverse the tail and \cref{enm:compreq:finalstep} to drop the final three robots at the tail's end.
This final drop is slightly more involved: if the last three coordinates form, e.g., a straight path of edge length $1$, our drawing formation cannot drop all robots at once.
With the help of \cref{enm:compreq:finalstep}, we handle this via an intermediate step.

\begin{lemma}%
\label{lem:drawing-body}
Consider a compatible drawing path $\bm{v} = \intoo{v_j}_{j=1}^{k}$ of a pattern $P$.
Let $\cR$ be the configuration formed by a drawing formation $F$ of size $\abs{P}$ in state $1$ anchored in $v_1$ that is compatible with $\bm{v}$.
Then $F$ can traverse $\bm{v}$ by taking one edge per \LCM/-cycle while dropping one robot at each coordinate in $\covered(v_j)$ when leaving $v_j \not\in \tail(\bm{v})$.
\end{lemma}
\begin{proof}
    \textbf{Enumeration of States.}
    We defined in \cref{def:df-state-enum} an enumeration of $A^\ell_F$, let the $i$-th state of $A^\ell_F$ be $state(i, \ell)$.
    We define the following unique states for the path, using an enumeration that includes different sizes $\ell$ of drawing formations fitting to the number of not dropped robots at a node.
    $$f(v_i) := |\covered((v_j)_{j=i}^k)|$$
    $$g(v_i) := \max\left(|(v_j)_{j<i}^{i-1}|\right) \textrm{ with } \covered((v_j)_{j<i}^{i-1}) = \emptyset$$
    $$state(v_i) := state(g(v_i) + 1, f(v_i))$$

    From (3) and (4) of \cref{def:compatibiliy} follows directly, that all such states exist.
    
    \textbf{Induction Proof.}
    Assumption: $F$ with anchor on $v_i$ with $|F| = |\covered((v_j)_{j=i}^k)|$ in state $i$ and dropped robots on $\covered((v_j)_{j=1}^{i-1})$.
    Let $H_F = (a, \vec{d}, \phi, \Delta)$ be the drawing hull of $F$ (\cref{def:drawing-hull}).
    We define the coordinate system $\cS$ as the coordinate system with $x$ direction $\vec{d}$ and origin at $v_1$.
    Start: At node $v_1$ this is initially given.

    Step:
    No coordinates $p \in \cB(v_i, 1-\delta)$ contain robots $r \notin F$, otherwise $p \in \covered(v_j) \cap \cB(v_i, 1-\delta), j < i$ with is a contradiction to \cref{def:covered-coordinate}. 
    Therefore $F$ is valid (\cref{def:valid-df}). 
    $r \in F$ knows anchor $a$ and direction vector $\vec{d}$ of $F$ (see \cref{lem:valid-df}) and $F$ is unabigous because it is valid (i.e. $r \in F$ is not in another $\epsilon$-granular drawing formation).
    Because $v$ is a drawing path (\cref{def:drawing-path}), $r$ can compute $v$ from $P$.
    With the assumption that $a$ (the anchor of $F$) is at $v_i$ adn $F$ in state $state(v_i)$, it can determine the coordinate system $\cS$.
    $dist(v_i, v_{i+1}) \leq 1-\delta$ (by \cref{def:drawing-path}) and $1-\delta \leq 1-\Delta$ (by \cref{def:compatibiliy}).
    From \cref{obs:moving} follows that $F$ can move from $v_i$ to $v_{i+1}$.
    $F$ moves such that its new anchor is $v_{i+1}$, $|\covered(v_i)|$ robots are dropped onto the coordinates $\covered(v_i)$ and its new state $state(v_{i+1})$.

\end{proof}

\begin{lemma}[{name=, restate=[name=restated]lemDrawingTail}]%
\label{lem:drawing-tail}
Consider a compatible drawing path $\bm{v} = \intoo{v_j}_{j=1}^{k}$ of a pattern $P$.
Let $\cR$ be the configuration that has
\begin{enumerate}
\item one robot at each coordinate in $P \setminus \covered(\tail(\bm{v}))$ and
\item a drawing formation $F$ of size $3$ in state $\abs{\tail(\bm{v})}$ anchored in $v_k$ that is compatible with $\bm{v}$.
\end{enumerate}
Then $F$ can dissolve within two \LCM/-cycles while dropping one robot at each coordinate in $\covered(\tail(\bm{v}))$.
\end{lemma}
The proof of \cref{lem:drawing-tail} is given in \cref{sec:drawing-tail-proof}.

\subsection{Full Pattern via Many Drawing Formations}%
\label{sec:df-pattern}

As shown in \cref{sec:drawing-symmetric-component}, we can draw any pattern $P$ \emph{if} we start in a suitable drawing formation $F$ (and have a compatible drawing path).
But we must first form such a drawing formation from the initial near-gathering, which might have a symmetricity $s > 1$.
In that case, since any drawing formation has symmetricity $1$, we cannot form $F$ (by \cref{thm:symmetry-condition}).
Instead, we show how to form $\sym(P)$ symmetric copies of $F$ that are placed such that they
\begin{enumerate}
\item have symmetricity $\sym(P)$ (we have $s \mid \sym(P)$ or we cannot form $P$, even globally) and
\item do not interfere with each other (if using suitable drawing paths, see \cref{sec:dp-implementation}).
\end{enumerate}

We start by partitioning the pattern $P$ of symmetricity $s_p \coloneqq \sym(P)$ into $s_P$ \emph{symmetric components}, each of which will be drawn by its own drawing formation.
\begin{definition}[Cone \& Symmetric Component]%
\label{def:symmetric-component}
Let $\vec{e}_x \coloneqq (1, 0)$.
For a pattern $P$ of symmetricity $s_P$, define the $i$-th \emph{cone}
\begin{dmath}
C^{(i)}
\coloneqq
\set{p \in \R^2 | \angle(\vec{e}_x, p) \in \intco{(i - 1) \cdot 2\pi / s_P, i \cdot 2\pi / s_P}}
\end{dmath}
and the $i$-th \emph{symmetric component}
\begin{math}
P^{(i)}
\coloneqq
P \cap C^{(i)}
\end{math}.
\end{definition}

\begin{figure}
    \begin{subfigure}[t]{0.30\linewidth}
        \centering\includegraphics[page=7, width=0.9\linewidth]{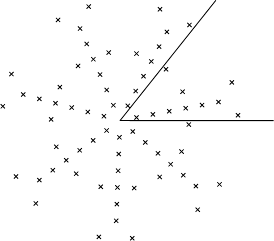}
        \caption{
            Pattern $P$ with symmetricity $s_P = 7$ and cone $C^\einindex$.
        }\label{fig:pattern-and-cone}
        \end{subfigure}
        \hfill
    \begin{subfigure}[t]{0.34\linewidth}
    \centering\includegraphics[page=8, width=0.8\linewidth]{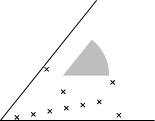}
    \caption{
        Symmetric component $P^{(1)}$ with an aligned drawing formation $F^{(1)}$.
    }\label{fig:sym-comp-and-df}
    \end{subfigure}
    \hfill
    \begin{subfigure}[t]{0.30\linewidth}
    \centering\includegraphics[page=9, width=0.9\linewidth]{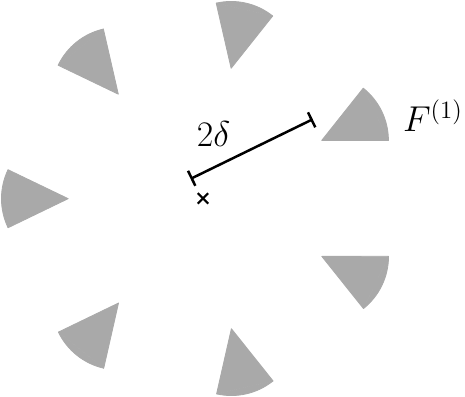}
    \caption{
        Initial drawing pattern $\cI$ for $s_P = 7$.
    }\label{fig:initial-drawing-formations}
    \end{subfigure}
    \caption{}
\end{figure}

Note that the symmetric components are pairwise disjoint and that $P = \bigcup_{i = 1}^{s_P} P^{(i)}$.
See \cref{fig:pattern-and-cone} for an illustration.

To form pattern $P$, we first form a suitable \emph{initial drawing pattern} of diameter $\leq 1$.
This initial drawing pattern places each robot $r \in \cR$ in one of $s_P$ $\epsilon$-granular drawing formations $F^{(i)}$ of size $\abs{P}/s_P$ in state $1$.
If there exists a drawing path $\bm{v}^{(1)}$ for $P^{(1)}$ that is compatible with $F^{(1)}$ and starts at the anchor of $F^{(1)}$, we immediately get a corresponding (rotation-symmetric) drawing path for each $F^{(i)}$.
Assuming that, additionally, those drawing paths lie \enquote{sufficiently inside} their respective cone $C^{(i)}$, we will prove a generalization of \cref{lem:drawing-body, lem:drawing-tail}, basically showing that the different $F^{(i)}$ can draw their $P^{(i)}$ without interfering with each other.
The existence of suitable drawing paths is shown in \cref{sec:dp-implementation}.

To enable the robots to deduce the symmetric component $P^{(i)}$ they are drawing\footnote{
    Since robots are disoriented, they cannot deduce which $P^{(i)}$ they are drawing.
    But they can deduce $P^{(i)}$'s coordinates in their own, local coordinate systems.
}, we require $F^{(i)}$ to be \emph{aligned} with $P^{(i)}$:
\begin{definition}%
\label{def:aligned_df}
Fix $i \in \set{1, 2, \dots, s_P}$ and the $i$-th symmetric component $P^{(i)}$ of a pattern $P$ with symmetricity $s_P$.
Let $F$ be a drawing formation with direction $\vec{d}$ and span $\phi$.
Then $F$ is \emph{aligned} with $P^{(i)}$ if $\angle(\vec{d}, \vec{e}_x) = (i - 1) \cdot 2\pi / s_P$ and if $\phi = \min\set{2\pi/s_P, \pi/3}$.
\end{definition}
\Cref{fig:sym-comp-and-df} gives an illustration.
With this, we define the \emph{initial drawing pattern} of a pattern $P$ and a suitable drawing formation $F$ as follows (illustrated in \cref{fig:initial-drawing-formations}):
\begin{definition}[Initial Drawing Pattern]%
\label{def:drawing_formation_pattern}
\label{def:initial-drawing-formation}
Fix a pattern $P$ of symmetricity $s_P$.
An \emph{initial drawing pattern} $\cI$ for $P$ is a configuration of $\abs{P}$ robots that consists of $s_P$ drawing formations $\set{F^{(i)}}_{i=1}^{s_P}$ of diameter $\Delta \leq 1/6$ in state $1$ such that:
\begin{enumerate}
\item $F^{(1)}$ is aligned with $P^{(1)}$ and anchored in $(2\Delta, \pi/s_P)$.
\item $F^{(i)}$ is a rotation of $F^{(1)}$ by $(i-1) \cdot 2\pi/s_P$.
\end{enumerate}
We say $\cI$ is $\epsilon$-granular if the $F^{(i)}$ are $\epsilon$-granular.
\end{definition}
Note that, by construction, each $F^{(i)}$ is aligned with $P^{(i)}$.
Moreover, $\cI$ is a near-gathering configuration (i.e., has diameter $\leq 1$) and has symmetricity $s_P$, such that we can form $\cI$ from any near-gathering for which the symmetry condition (\cref{thm:symmetry-condition}) holds.

It remains to prove that once the initial drawing pattern is formed, each drawing formation $F^{(i)}$ forms its symmetric component $P^{(i)}$ and does not interfere with the operation of any other drawing formation $F^{(j)}$ with $j \neq i$.
\begin{lemma}%
\label{lem:drawing-full-pattern}
Assume the current configuration is an $\epsilon$-granular initial drawing pattern $\cI$ for a pattern $P$ of symmetricity $s_P$.
Consider a drawing path $\bm{v}^{(1)} = \intoo[\big]{v^{(1)}_j}_{j=1}^{k}$ of symmetric component $P^{(1)}$ that is compatible with $F^{(1)}$ such that
\begin{enumerate}
\item the path $\bm{v}^{(1)}$ starts in the anchor of $F^{(1)}$,
\item $\bm{v}^{(1)}$ lies in the first cone (i.e., $\bm{v}^{(1)} \subseteq C^{(1)}$) and
    \begin{dmath}
    \dist\intoo*{\bm{v}^{(1)}, \partial C^{(1)}}
    >
    \max\Set{\epsilon, \Delta \cdot \sin\intoo*{\nicefrac{2\pi}{s_P}}}
    \end{dmath}.
\end{enumerate}
Then $P$ can be formed in $\LdauOmicron{k}$ many \LCM/-cycles.
\end{lemma}
\begin{proof}

    From \cref{lem:drawing-body,lem:drawing-tail} we know, $P^\einindex$ can be formed by $F^\einindex$ assuming $P^\einindex$ is the whole pattern.
    The traversal of this path takes at most $k$ rounds to reach $v^\einindex_k$ plus 2 rounds for dropping the last robots.
    With the other symmetric components $P^\iindex$ next to $P^\einindex$, $F^\einindex$ can still form $P^\einindex$ if $F^\einindex$ is always valid.
    We will prove in the following, that $F^\einindex$ is always valid.

    \textbf{Validity}.
    $F^\einindex = (\cR_F, H_F)$ is valid, if there exists no subset $\cR_G \subseteq \cR$ which fulfills the criteria of \cref{def:epsilon-granular-df} such that $G = (\cR_G, H_G)$ a $\epsilon$-granular drawing formation $\cR_G \neq \cR_F$ and $H_F \cap H_G \neq \emptyset$.
    $F^\einindex$ is aligned with $C^\einindex$ (\cref{def:aligned_df}).
    It follows directly, that $F^\einindex \subseteq C^\einindex$ and naturally for all symmetric drawing formation $F^\iindex$ that $F^\iindex \subseteq C^\iindex$.
    Therefore, those drawing formations have disjunct hulls.
    From the proof of \cref{lem:drawing-body} we know that all dropped robots on $p \in P^\einindex$ have a distance $\geq \Delta$ to $H_F$, therefore those robots cannot be part of $\cR_G$.
    It is left to show, that $r \in C^\iindex$ with $i \neq 1$ cannot build a drawing formation $G$ that intersects $H_F$.
    
    Two dropped robots $r_1, r_2$ on $p_1, p_2 \in P$ have $dist(r_1, r_2) > \epsilon$, this follows from the fact that $\varepsilon \leq \mindist(P)$ (\cref{enm:compreq:params} of \cref{def:compatibiliy}).
    Let $F$ be anchored on $v^\einindex_j$.
    By prerequisit (1) $dist(v^\einindex_j, \partial \C^\einindex) > \epsilon$.
    Because $F^\einindex$ is aligned to $C^\einindex$, $dist(\partial C^\einindex, \cR_F) > \epsilon$.
    Therefore $r_1 \in F^\einindex$ and $r_2 \notin F^\einindex$ have $dist(r_1, r_2) > \epsilon$.
    Therefore, drawing formation $G$ must have a pair of $dist(r_1, r_2) > \epsilon$ which are part of the same drawing formation $F^\iindex$.
    There must exist a third robot $r_3 \notin F^\iindex$ collinear to $r_1, r_2$ with $dist(r_3, \{r_1, r_2\}) \leq \Delta$. 
    W.l.o.g $i = 1$.
    Let $H_F = (a, \vec{d}, \Phi, \Delta)$ be the drawing hull of $F^\einindex$.
    Because $F^\einindex$ is aligned to $C^\einindex$ (\cref{def:aligned_df}) $\vec{d}$ is parallel to one side of its boundary and the line segment $\geomline(a, -\vec{d}, \Delta)$ can never cut this side.
    $\geomline(a, -\vec{d}, \Delta - \epsilon)$ cuts the other side of $C^\einindex$ boundary with angle $\Phi$.
    The length of $\geomline(a, -\vec{d}, \Delta - \epsilon)$ must be $\geq \frac{dist(\partial C^\einindex)}{\sin(\Phi)}$.
    The line segment has a length of $\Delta$.
    $\Phi = \frac{2\pi}{s_P}$ (\cref{def:aligned_df})
    $dist(a, \partial C^\einindex) \geq \Delta \cdot \sin\intoo*{\frac{2\pi}{s_P}}$ (assumption (2) of this lemma).
    This resolves to $\Delta - \epsilon \geq \Delta \cdot \sin\intoo*{\frac{2\pi}{s_P}} / \sin\intoo*{\frac{2\pi}{s_P}} = \Delta$, which is obviously a contradiction.
    Therefore, $\geomline(a, -\vec{d}, \Delta - \epsilon)$ is completely in $C^\einindex$ and cannot contain $r_3 \notin F^\einindex$.
\end{proof}

\subsection{Putting Everything Together}%
\label{sec:main-result-proof}

\Cref{sec:df-pattern} showed how to form a pattern $P$ assuming we start in a suitable initial drawing pattern $\cI$ and if a compatible drawing path $\bm{v}^{(1)}$ for $P^{(1)}$ exists.
We continue by showing the existence of such a path for patterns with symmetricity $s_P \coloneqq \sym(P) < \abs{P} / 2$ (\cref{lem:existence-of-suitable-drawing-path}, proven in \cref{sec:dp-implementation,sec:proof-existance-path}); patterns of larger symmetricity can be handled without drawing formations (\cref{lem:star-pattern}).
Afterward, in \cref{lem:phase-distinction}, we prove that each robot can distinguish in which phase of our protocol it is:
\begin{enumerate*}[(i)]
\item forming $\cI$,
\item being part of a valid $\epsilon$-granular drawing formation $F$,
\item dropping the last three robots at the tail's end, or
\item having been dropped at a pattern coordinate.
\end{enumerate*}
Putting everything together, we conclude this \lcnamecref{sec:general_approach} with the proof of \cref{thm:main-result}.

\begin{lemma}[{name=, restate=[name=restated]lemExistDrawingPath}]%
\label{lem:existence-of-suitable-drawing-path}
Consider the $\epsilon$-granular drawing formation $F^{(1)}$ of the initial drawing pattern for a connected pattern $P$ of symmetricity $s_P < \abs{P}/2$.
The parameter $\epsilon$ can be chosen such that $F^{(1)}$ has $\abs{L_{F^{(1)}}(\epsilon)} \geq 2 + \abs{P^{(1)}}$ $\epsilon$-granular locations.
Moreover, there exists a drawing path $\bm{v}^\einindex$ of symmetric component $P^{(1)}$ that is compatible with $F^{(1)}$ such that
\begin{enumerate}
\item the path $\bm{v}^{(1)}$ starts in the anchor $a = (2\Delta, \pi/s_P)$ of $F^{(1)}$,
\item $\bm{v}^{(1)}$ lies in the first cone (i.e., $\bm{v}^{(1)} \subseteq C^{(1)}$) and
    \begin{dmath}
    \dist\intoo*{\bm{v}^{(1)}, \partial C^{(1)}}
    >
    \max\Set{\epsilon, \Delta \cdot \sin\intoo*{\nicefrac{2\pi}{s_P}}}
    \end{dmath}.
\end{enumerate}
\end{lemma}

\begin{lemma}[{name=, restate=[name=restated]lemStarPattern}]%
\label{lem:star-pattern}
Consider a pattern $P$ with symmetricity $s_P \geq |P|/2$.
$P$ can be formed.
\end{lemma}

With a separate algorithm, we can form patterns with symmetricity $n$ or $\nicefrac{n}{2}$.
The robots form $P$ scaled down to diameter $1$.
Then, the robots scale the small pattern back to its original (large) size.
Because of the high symmetricity, the scaling can be performed locally.
The proof can be found in \cref{sec:star-pattern}.

\begin{lemma}%
\label{lem:phase-distinction}
Let $r \in \cR$ be a robot in a configuration described in \cref{lem:existence-of-suitable-drawing-path} executing the protocol from \cref{lem:drawing-full-pattern}.
Then $r$ can locally distinguish between the following situations:
\begin{enumerate}[(i)]
\item\label{itm:lem:phase-distinction:formingI} $r$ is in an initial configuration before the initial drawing pattern is formed
\item\label{itm:lem:phase-distinction:drawing} $r \in H_F$ of a valid $\epsilon$-granular drawing formation $F$
\item\label{itm:lem:phase-distinction:taildropping} $r \in F_{inter}$ (see \cref{def:intermediate-shape})\footnote{$F_{inter}$ is a slight alteration of the $\epsilon$-granular drawing formation used in the very last step of the execution.}
\item\label{itm:lem:phase-distinction:dropped} $r$ has been dropped from a drawing formation
\end{enumerate}
\end{lemma}
\begin{proof}
\begin{enumerate}[(i), wide=0pt]
\item
    If a robot $r \in \cR$ sees $\abs{P}$ robots, it is in a near-gathering and observes all robots $\cR$.
    In that case, $r$ checks whether $\cR$ equals the initial drawing pattern $\cI$ (\cref{def:initial-drawing-formation}) or any of the execution steps resulting from the protocol described in \cref{lem:drawing-full-pattern} that are still a near-gathering ($\ldauTheta{1}$ many).
    If not, $r$ can conclude that it is in the initial configuration.

\item Is clear by \cref{lem:valid-df}.

\item Is shown in \cref{lem:intermediate-shape-valid}.

\item
    When a robot is dropped from a drawing formation it is on $p \in P$.
    We know, that $\mindist(P) > \epsilon$, therefore dropped robots can never form an $\epsilon$-granular drawing formation.
    Moreover, in \cref{lem:drawing-full-pattern} we have shown that all $\epsilon$-granular drawing formations in the configuration are valid.
    Therefore, no robot $r' \in H_F$ can be part of another $\epsilon$-granular drawing formation $F'$.
    Similar arguments are true for $F_{inter}$.
    Hence, a robot observing that it is not in one of the first three situations knows that it has been dropped.
    \qedhere
\end{enumerate}
\end{proof}

\thmMainResult*
\begin{proof}
The first direction follows from \cref{thm:symmetry-condition}, since a pattern where $s_I \coloneqq \sym(I)$ does not divide $s_P \coloneqq \sym(P)$ cannot be formed.

For the second direction, assume $s_I \mid s_P$.
By \cref{lem:star-pattern} the pattern can be formed if $s_P \geq \abs{P}/2$, so assume $s_P < \abs{P}/2$.
Then we must execute the protocol described in \cref{lem:drawing-full-pattern}, whose prerequisites (esp.~the existence of a suitable drawing path) can be fulfilled:
By \cref{lem:phase-distinction} we know, that a robot can locally decide between the phases necessary to start and execute the protocol.
If $\cR$ is in an initial near-gathering before forming the initial drawing pattern $\cI$, the robots collectively form $\cI$ (which has symmetricity $s_P$ and, thus, can be formed by \cref{thm:symmetry-condition}).
From \cref{lem:existence-of-suitable-drawing-path}'s guarantee on $\abs{L_{F^{(1)}}(\epsilon)}$ we get that the drawing formation has enough locations for the number of robots as well as the existence of a suitable drawing path.
Thus, we can apply \cref{lem:drawing-full-pattern} to get that one robot is dropped at each $p \in P$ after at most $\ldauOmicron{\abs{P}}$ rounds.
By \cref{lem:phase-distinction}, robots can realize that they have been dropped and remain idle on their respective pattern coordinate.
\end{proof}
}
{\newcommand{\tfinal}{\ensuremath{t_{final}}}
\newcommand{\tprefinal}{\ensuremath{t_{prefinal}}}

\section{Existence of Suitable Drawing Paths}%
\label{sec:dp-implementation}

In the previous section, we showed that a drawing formation $F^\iindex$ which is placed in $C^\iindex$ can traverse a drawing path $\bm{v}$ of $P^\iindex$ if $F^\iindex$ is compatibiliy to $\bm{v}^\iindex$.
We omited all details on how we can create such a drawing path.
In this section, we will construct a path and prove that it fulfills all required properties of \Cref{lem:existence-of-suitable-drawing-path}.
The final proof of \Cref{lem:existence-of-suitable-drawing-path} can be found in \Cref{sec:proof-existance-path}.

\paragraph{Outline}
For a symmetric component $P^\einindex$, we define a tree $T^\einindex$ with $\cO(n)$ nodes such that its nodes cover all points of $P^\einindex$.
The tree's root node will be $(2\Delta, \pi/s_P)$, the initial position of the drawing formation $F^\einindex$ aligned to $P^\einindex$.
It is clear that a simple traversal of $T^\einindex$ will fulfill the requirement (1) and (4) of compatibility.
To additionally fulfill (2) and (3), we construct a tail that fits the requirements and append it to the traversal.
To prove that such a tail always exists we show, that it is always possible to rotate $P$ s.t. $P^\einindex$ contains $\geq 3$ connected positions.\footnote{While the pattern is connected by definition, the cuts in symmetric components can disconnect parts of the component. E.g. if the pattern is a multi-helix spiral}
We use these three positions to append a tail to $T^\einindex$ that fulfills the requirement of compatibility.

\subsection{Tree Construction} 
\label{ssec:building-path-system}
\begin{definition}[Drawing-Tree]
	\label{def:drawing-tree}
	Let $P^\einindex$ be a symmetric component of $P$.
	We call $T^\einindex$ constructed with algorithm \cref{alg:compute-drawing-tree} a \emph{drawing-tree} of $P^\einindex$.
\end{definition}

\begin{algorithm}
	\caption{\textsc{ConstructDrawingTree}($P^\einindex$)}
		\label{alg:compute-drawing-tree}
	\begin{algorithmic}
		\State $root \gets (2\Delta, \pi/s_P)$
		\State $Base_{left} \gets$ linear line $(2\Delta, \pi/s_P) + i \cdot(4\delta, 2\pi/s_P)$ for $i \in \{1, \cdots, \infty\}$ starting at $root$
		\State $Base_{right} \gets$ linear line $(2\Delta, \pi/s_P) + i \cdot(4\delta, 0)$ for $i \in \{1, \cdots, \infty\}$ starting at $root$
		\State $T^\einindex\gets root + Base_{left} + Base_{right}$ \Comment{Base Tree}
		\While{$cov(T^\einindex) \neq P^\einindex$} \Comment grow the tree inside the cone
		\State let $(p, t)$, $p \in P^\einindex \setminus cov(T^\einindex)$, $t \in T^\einindex$ be the pair with minimal distance
		\If{$dist(p, t) < 1-\delta$}
		\State add $p$ to $T^\einindex$ and connect it to $t$
		\Else
		\State $t' \gets (p + t)/2$ \Comment Intermediate node between $p$ and $t$
		\State add $t'$ to $T^\einindex$ and connect it to $t$; add $p$ to $T^\einindex$ and connect it to $t'$
		\EndIf
		\EndWhile
		\State remove all subtrees of $T^\einindex$ that do not cover any $p \in P^\einindex$ and \Return $T^\einindex$
	\end{algorithmic}
\end{algorithm}

It is clear that $T^\einindex$ is computable from $P$.
The nodes cover all positions in $P^\einindex$. 
Distances between neighboring nodes are $4\delta$ on the linear lines and at most $1- \delta$ everywhere else.
So $T^\einindex$ fulfills all requirements for a drawing path for $\delta \leq 0.2$.\footnote{It holds for $\delta \leq 0.2$ that $4\delta \leq 1 - \delta$}

\begin{observation}
	\label{obs:drawing-tree-is-path}
	A reasonable short and deterministically computable traversal of $T^\einindex$ as defined in \Cref{def:drawing-tree} is a \emph{drawing path} for $\delta \leq 0.2$. 
	We define $trav(T^\einindex)$ to represent the path of this traversal.
\end{observation}

\paragraph{Summary of \cref{lem:existence-of-suitable-drawing-path} proof}
$trav(T^\einindex)$ is a drawing path, but it is not necessarily compatible to a drawing formation because the tail does always not fulfill the requirements of \Cref{def:compatibiliy}.
The tail is the last part of the path, which covers in total $\leq 3$ pattern positions (\cref{def:dp-tail}).
For the compatibility, it must have a length that is traversable by a drawing formation of $3$ robots, i.e. length $\leq \abs{A_F^{3}(\epsilon)}$.
Additionally, all positions covered by the tail must be in the distance $\leq 1$ to the last node of the tail.
This allows the remaining $3$ robots of the drawing formation to reach the last three pattern positions from the last node of the tail.
In \cref{lem:suitable-tail-exists} we first show that there exists suitable start point $z_{start}$ and end point $z_{end}$ for such a tail. 
$z_{end}$ is a suitable end point, when it has at least $3$ positions $p_1, p_2, p_3\in P^\einindex$ in its $1$-surrounding.
$z_{start}$ must have a constant distance to $z_{end}$ and cover at least one additional position $p_4 \in P^\einindex \setminus \set{p_1, p_2, p_3}$.
To prove \cref{lem:existence-of-suitable-drawing-path} we construct a compatible drawing path out of $z_{start}$, $z_{end}$ and $T^\einindex$.
We connect $z_{start}$ and $z_{end}$ with intermediate nodes in a straight line and add possibly up to 3 additional intermediate nodes around $z_{end}$ to cover the positions $p_1, p_2$ and $p_3$.
$z_{start}$ is connected with a straight line of intermediate nodes to the end of $T^\einindex$ as well.
Because $z_{start}$ and $z_{end}$ have a constant distance and cover, together with the intermediate nodes, at least the positions $p_1, \cdots, p_4$ we know that the tail fulfills the requirements mentioned above.
The full proof of \cref{lem:existence-of-suitable-drawing-path} can be found in \cref{sec:proof-existance-path}.

}
{\section{Discussion and Future Work}%
\label{sec:conclusion}

This \lcnamecref{sec:conclusion} discusses some additional aspects of our arbitrary pattern formation protocol for oblivious robots with a limited viewing range and highlights some open questions.

\paragraph{Near-Gathering with Symmetry Preservation}
We presented a protocol that starts in a near-gathering (all robots within a constant diameter).
The authors of \cite{DBLP:conf/opodis/CastenowH0KKH22} gave a class of near-gathering protocols in the same model we consider.
They proved that their protocols, starting from any connected initial configuration, reach a near-gathering in $\cO(|P|^2)$ rounds.
However, their class contains protocols that increase the symmetricity during the execution.
For the application of pattern formation, it is essential that the swarms symmetricity does not exceed $\sym(P)$.
It remains an interesting open question whether there is a suitable near-gathering protocol that preserves the initial symmetricity.

\paragraph{Synchronicity}
We assume the fully-synchronous \fsync/ scheduler for our protocol.
In the related work, many papers assume \async/.
The authors of \cite{DBLP:journals/siamcomp/FujinagaYOKY15} proved that, for an unlimited viewing range, \fsync/ has the same pattern formation capabilities as \async/.
An unlimited viewing range makes it much easier to maintain common knowledge in the swarm (like a common coordinate system).
For a limited viewing range, our protocol must maintain this information during execution (using the $\epsilon$-granular drawing formations).
In the \async/ model, where only a part of the drawing formation might be activated, we would have to ensure that \enquote{partially} moving a drawing formation does not destroy the encoded information (e.g., by encoding information redundantly).
It is a crucial part of $\epsilon$-granular drawing formations that their robots can identify them, and we would have to maintain this property under partial movements.
While it seems challenging, a careful design might be able to solve this.

\paragraph{Connectedness}
In our main theorem, we assume that the unit disc graph of $P$ is connected.
This is a natural assumption for robots with limited visibility because they cannot interact beyond their viewing range.
Whenever such pattern formation is used in a real-world application, basically only connected patterns are meaningful (e.g., for creating an ad-hoc-network).
However, our protocol is capable of forming patterns with less connectivity.
It applies to any pattern $P$ where a compatible drawing path can be created.
When we translate \Cref{def:compatibiliy} to a pattern, we get the following condition:

Let $perm(P)$ be a permutation of $P$.
There exist $perm(P) = (p_i)_{i=1}^{k}$ such that
\begin{enumerate}
    \item $\dist(p_i, p_{i+1}) \leq \abs{A_F^{k-i}(\epsilon)}$ for $1 \leq i \leq k-2$
    \item $p_{k-2}, p_{k-1}$ and $p_k$ must have a smallest enclosing circle of radius $\leq 1$
\end{enumerate}
Condition (1) follows from (1) and (2) of \Cref{def:compatibiliy}, and condition (2) is necessary such that a drawing path can fulfill (4) of \Cref{def:compatibiliy}.

Besides the last three robots, the maximal distance between two pattern positions is dependent on $\abs{A_F^{k-i}(\epsilon)}$.
In the proof of \Cref{lem:df-state-size} we have shown that $\abs{A_F^{k-i}(\epsilon)} = \cO\left(\binom{\epsilon^{-1}}{k-i-3}\right)$.
We can choose $\epsilon$ freely to reach any distance necessary.
}

\bibliography{static-and-custom-copy-downloaded-bibliography.bib}

\appendix
{\section{Proof of \Cref{lem:existence-of-suitable-drawing-path}}
\label{sec:proof-existance-path}
\begin{lemma}[{name=, restate=[name=restated]lemExistTail}]
	\label{lem:suitable-tail-exists}
	Let $P$ be a pattern with symmetricity $s_P < |P|/2$. 
	There exist for each symmetric component $P^\einindex$ with cone $C^\einindex$ points $z_{start}$ and $z_{end}$ such that
	\begin{enumerate}
		\item $|\cB(z_{end}, 1)| \geq 3$
		\item if $\abs{P^\iindex} > 3$ 
		\begin{enumerate}
			\item $\abs{\cB(z_{start}, 1-\delta) \cup \cB(z_{end}, 1)} \geq 4$
			\item $dist(z_{start}, z_{end}) = \cO(1)$
			\item $dist(\{z_{start}, z_{end}\}, \partial C^\einindex) = \Omega(1/s_P + \mindist(P))$
		\end{enumerate}
	\end{enumerate} 
\end{lemma}

\begin{proof}
	\textbf{(1)} 
	Consider a finite subset $S \subset \cR^2$ of symmetricity $s$ and size $\abs{S} \geq 3s$.
	Assume the unit disc graph $\cU(S)$ is connected.
	It is a simple geometric fact that there exists a subset $C \subseteq S$ of size $\abs{C} = 3$ and with $\angle(C) < 2\pi / s$ such that $\cU(C)$ is connected.
	We stated and proved this in the appendix (\Cref{lem:3connectedsectors}).
	Our pattern $P$ is such a set and a symmetric component $P^\einindex$ is a subset with $\angle(C) < 2\pi / s$, therefore there exists a rotation of $P$ such that $p_1, p_2, p_3 \in P^\einindex$ with $\cU(\{p_1, p_2, p_3\})$ connected.
	W.l.o.g.\ we assume this is the rotation of $P$.
	Then, there exists a point $z_{end}$ with $\cB(z_{end}, 1) \supseteq \{p_1, p_2, p_3\}$.

	\textbf{Trivial cases for (a), (b) and (c)}. 
    If $P^\einindex$ contains $4$ positions with $\cU(\{p_1, p_2, p_3, p_4\})$ connected, it is clear that $z_{start}$ and $z_{end}$ can be placed with 
	$\cB(z_{start}, 1-\delta) \cup \cB(z_{end}, 1) \supseteq \{p_1, p_2, p_3, p_4\}$ (a).
	If $|\cB(z_{end}, 1-\delta)| \leq 3$ (w.l.o.g. $p_4 \notin \cB(z_{end}, 1-\delta)$), we place $z_{start}$ in distance $\leq 1-\delta$ to $p_4$ (b).
	$z_{start}$ and $z_{end}$ have a constant distance (c).
    If $\cB(z_{end}, 1)$ contains more than three positions that are not all connected, the placement for $z_{start}$ is analog.

    \textbf{(a) and (b)} 
    We assume that $\cB(z_{end}, 1) = \{p_1, p_2, p_3\}$.
    Because $\cU(P)$ is connected, there exists a path from $p_1$ to $p' \in P^\einindex \setminus \{p_1, p_2, p_3\}$ in $\cU(P)$.
	Let $p_1, w_1 \cdots, w_k, p'$ be a shortest path form $p_1$ to $p'$.
	The path can only contain $3$ consecutive nodes in one symmetric component (otherwise the component would contain 4 connected positions), therefore there exist $w_j \in P^{(2)}$ with $j \leq 3$ and $w_l \notin P^{(2)}$ with $l \leq j + 3$.
	If $w_l \in P^\einindex,$ we find $p_4 = w_l$ with $dist(p_1,p_3) \leq 6$.
	If $w_l \in P^{(3)}$, there exist a rotational symmetric point in $P^\einindex$, let this be $p_4$.
	It is clear that $dist(p_1, p_4) \leq dist(p_1, w_l) \leq 6$.
	We can place $z_{start}$ in the distance $1-\delta$ of $p_4$ to fulfill (a) and (b).

	\textbf{(c)}
	$z_{start}$ can be placed relatively freely in a radius of $1-\delta$ around $p_4$, this easily fulfill (c).
	There exist cases where $z_{end}$ must be placed directly on one of the three connected positions, let this be $p_1$.
	From \Cref{lem:3connectedsectors} we can follow, that w.l.o.g.\ $p_{end}$ lies on the bisector of $C^\einindex$.
	Let $p' P^{(2)}$ be the rotational symmetric point to $p_1$. 
	Naturally, $dist(z_{end}, \partial C^\einindex) = \frac{1}{2} dist(p_1, p') \geq mindist(P)$.
\end{proof}

\lemExistDrawingPath*

\begin{proof}

		Let $T^\einindex$ be the drawing tree of $P^\einindex$ \Cref{def:drawing-tree}.
		$trav(T^\einindex)$ has all properties for a drawing path (\Cref{obs:drawing-tree-is-path}).
		Let $F^\einindex$ be a drawing formation with $\epsilon = \Theta(\min(1/s_P, \mindist(P), 1/\sqrt{|P|})$ and $\Delta = 0.1$ and $\Phi = 2\pi/s_P$
		To make $trav(T^\einindex)$ compatible with $F^\einindex$ we append the points $z_{start}$ and $z_{end}$ from \Cref{lem:suitable-tail-exists}.
		We add $b=\cO(P)$ itermediate nodes $w_{1}, \cdots, w_{b}$ between the end of $trav(T^\einindex)$ and $z_{start}$. 
		We add $b' = \cO(1)$ intermediate nodes $w_{b+1}, \cdots, w_{b+b'}$ between $z_{start}$ and $z_{end}$
		The resulting path is
		$$\bm{v}^\einindex := trav(T^\einindex) + (w_i)_{i=0}^b + (z_{start}) + (w_i)_{i={b+1}}^{b+b'} + (z_{end})$$
		We make sure, that 
		$$\covered((w_i)_{i={b+j}}^{b+b'} + (z_{end})) \subseteq \cB(z_{end}, 1) \textrm{ with } |\covered((w_i)_{i={b+j}}^{b+b'} + (z_{end}))| \geq 3$$ 
		for $1 \leq j \leq b'$. 
		This is possible by placing up to $3$ intermediate nodes in distance $\leq \delta$ to $z_{end}$.
		This number of nodes is sufficient to reach $z_{start}$, respectively $z_{end}$, with distances $\leq 1-\delta$ between $w_i$ and $w_{i+1}$, because $z_{start}$ has a distance $\cO(|P|)$ from any node of $trav(T^\einindex)$ and $z_{end}$ has a constant distance from $z_{start}$ (see \Cref{lem:suitable-tail-exists}).
		There obviously exist deterministic methods to define the intermediate paths, chose $z_{end}$, $z_{start}$, and determine $trav(T^\einindex)$ with $hops(trav(T^\einindex)) = \cO(|P|)$.

	\textbf{Compatibiliy}
		We show that conditions (1) - (4) from \Cref{def:compatibiliy} are fullfiled.
		\textbf{(1)} with $\delta = 0.1$ this is fullfiled
		\textbf{(2)}
		From \Cref{lem:suitable-tail-exists} we know that $|\cB(z_{end})| = 3$.
		From the equation in the beginning of this proof follows $\covered(tail(v^\einindex)) \subseteq \cB(z_{end})$,
		\textbf{(3)}
		If $|P^\einindex| \geq 4$ than $\abs{\cB(z_{start}, 1-\delta) \cup \cB(z_{end}, 1)} \geq 4$ (by \Cref{lem:suitable-tail-exists} (2))
		Hence, the tail of $v^\einindex$ must start after $z_{start}$.
		$|tail(v^\einindex)| = \cO(1)$ in this case (3) is fulfilled (see \Cref{lem:df-state-size}).
		If $|P^\einindex| = 3$ we know that $\epsilon = \cO(1/|P|)$.
		By \Cref{lem:df-state-size}) follows that $\cA^3_F(\epsilon) = \Omega(|P|)$.
		$tail(v^\einindex) = v^\einindex$ with length $\cO(|P|)$ in this case. 
		This fulfills (3).
		\textbf{(4):}
		With $\epsilon = \cO(1/\sqrt{|P|})$ we have $|A^4_F(\epsilon)| = \Omega(|P|)$ (see \Cref{lem:df-state-size}).
		We have $|v^\einindex| = \cO(|P|)$. 
		This fulfills (4)

	\textbf{Requirements (1) and (2) of \Cref{lem:existence-of-suitable-drawing-path}.}
	(1) the root node of $T^\einindex$ has coordinate $(2\Delta, \pi/s_P)$ and is the start of $v^\einindex$.
	(2) By construction of \Cref{def:drawing-tree} is clear that $dist(t, \partial c\einindex) \geq dist((2\Delta, \pi/s_P), \partial c\einindex) = \Delta \cdot \sin \intoo[\big]{\frac{2\pi}{s_P}}$ for $t \in T^\einindex$.
	By \Cref{lem:drawing-tail} we know that 
	$dist(z_i, \partial C^\einindex) = \Omega(\mindist(P)), i \in \{1,2\}$.
	$\epsilon < \min(1/s_P, \mindist(P), 1/\sqrt{|P|}) \cdot \Delta$.
	With this choice of $\epsilon$ we can create a $\epsilon$-granular drawing formation that has more locations that $|P^\einindex|+2$.

	Such that $F$ is an $\epsilon$-granular drawing formation, the parameter $\Delta, \phi$ and $\epsilon$ must be known. 
	$\Delta$ and $\Phi$ are given above and computable from $P$.
	$\epsilon = \min(1/s_P, \mindist(P), 1/\sqrt{|P|}) \cdot c$ for a constant $c<1$. 
	The constant can be deterministically determined (but we never write it down).
\end{proof}
}
{\section{Proof of \Cref{lem:drawing-tail}}%
\label{sec:drawing-tail-proof}

\lemDrawingTail*

We have shown in \Cref{lem:drawing-body} that on all coordinates outside the $\tail(v)$, $F$ can drop robots while traversing the drawing path $v$.
On the tail it cannot further drop robots before reaching the end; otherwise $|\cR_F| \leq 2$, which is not an $\epsilon$-granular drawing formation anymore.
In fact, it can never be a valid drawing formation because two robots can not encode the direction $\vec{d}$ in a model without a compass.
Therefore, $F$ does not drop robots onto $\covered(tail(v))$ during the traversal.
Instead, the drawing formation moves onto $v_k$, and the robots will move from there onto $\covered(tail(v))$.
Because $dist(\covered(tail(v)), v_k) \leq 1$ for a compatible path, the drawing formation is close enough to all remaining positions.
However, not all robots in $\cR_F$ are on $v_k$; they can have a distance up to $\Delta$.
If $dist(v_k, p) > 1 - \Delta, p \in \covered(tail(v))$ than the drawing formation $F$ can be placed inconvinenetly such that $dist(r, p) > 1, r \in \cR_F$.
We will add an intermediate step that reshapes the drawing formation such that all robots have a distance of $\leq 1$ to the coordinate they must obtain in the end.
Robots may leave the drawing hull $H_F$, but most properties of a drawing formation must still be fulfilled.
This intermediate shape $F_{inter}$ must be valid in the sense that robots in $F_{inter}$ must be able to determine the position of $v_k$ and the direction vector $\vec{d}$ to compute the global coordinate system.
In the following proof, we will define the intermediate shape and prove that robots can obtain this information.

\begin{definition}[$\epsilon$-intermediate-shape]
	\label{def:intermediate-shape}
	Let $P$ be a pattern and $v = (v_1, \cdots, v_k)$ be its drawing path with 
	\begin{enumerate}
		\item $|\covered(tail(v))| = 3$ and
		\item $\cB(v_k, 1) \supseteq \covered(tail(v))$
	\end{enumerate}
	Let $\covered(tail(v) = \{p_1, p_2, p_3\}$.
	The intermediate shape $F_{inter}$ definines the following positions for a set of three robots $r_1, r_2, r_3$.
	\begin{itemize}
		\item $r_1$ is on $v_k$
		\item $r_2$ is distance $\epsilon/2$ in direction $p_2$ and 
		\item $r_3$ is distance $\epsilon/3$ in direction $p_3$
	\end{itemize}
\end{definition}

\begin{observation}
	\label{obs:intermediate-shape-distance}
	Let $F_{inter}$ be an intermediate shape as in \Cref{def:intermediate-shape}.
	It is clear, that $dist(r_1, p_1) \leq 1$, $dist(r_2, p_2) \leq 1$ and $dist(r_3, p_3) \leq 1$.
\end{observation}

\begin{lemma}
	\label{lem:intermediate-shape-valid}
	Let $P$ be a pattern and $v = (v_1, \cdots, v_k)$ be a drawing path which is compatible with an $\epsilon$-ganular drawing formation.
	Let $F_{inter}$ be an intermediate shape as defined in \Cref{def:intermediate-shape} which is on $v_k$.
	Let $dist(v_k, \cR \setminus F_{inter}) > \epsilon$. 
	All robots $r \in F_{inter}$ can decide, that they are in $F_{inter}$ and can compute the positions of $tail(v_k)$ in their local coordinate system.
\end{lemma}

\begin{proof}
        To decide, whether a robot $r \in F_{inter}$ is in $F_{inter}$ it observes other robots in distance $\epsilon$.
        Because the $dist(v_k, \cR \setminus F_{inter}) > \epsilon$ it only finds one triple of robots with distances $\epsilon/2$ and $\epsilon/3$.
        Let $r_1, r_2, r_3, p_1, p_2, p_3$ be as in \Cref{def:intermediate-shape}.
        The triangle $r_1, r_2, r_3$ is never equiliteral (one side has length $\epsilon/2$ and another $\epsilon/3$).
        With chirality, all three robots know, who is $r_1, r_2$ and $r_3$.
        They know the coordinates $p_1, p_2$ and $p_3$ as well as the position of $v_k$ in the global coordinate system.
        The positions of $r_1, r_2$ and $r_3$ in the global coordinate system are also defined.
        Therefore, they can translate/rotate their local coordinate system and compute $tail(v_k)$.
\end{proof}

\Cref{lem:drawing-tail} follows immediately from \Cref{obs:intermediate-shape-distance} and \Cref{lem:intermediate-shape-valid}.
}

{\section{Proof of \Cref{lem:star-pattern}}
\label{sec:star-pattern}

\lemStarPattern*

In the following proof, we give an algorithm for patterns with a symmetricity of $n$ or $\nicefrac{n}{2}$.
If the diameter of $P$ is $\leq 1$, the pattern can be formed in one step, assuming suitable symmetricity.
Otherwise, the robots form $P$ scaled down to diameter $1$.
Then, the robots scale the small pattern back to its original (large) size.
We show that this algorithm can be computed and executed by the individual robots in our model with limited visibility and without memory.

\begin{proof}

		For $s_p = |P|/2$, the pattern $P$ has coordinates $\cup_{i=0}^{s_P-1}\{(D_1, \alpha_1 + i\cdot 2\pi/s_P), (D_2, \alpha_2 + i\cdot 2\pi/s_P)\}$ with $\alpha_1, \alpha_2 \leq  2\pi/s_P$. 
		$s_p = |P|$ is just a special case with $D_1 = D_2$ and $\alpha_1 = \alpha_2$.
		Let $pos_t(r_i)$ be the positon of the robot $r_i$ in round $t$.
		In a configuration with symmetricity $|P|/2$, the positions are as follows $pos_t(r_i) = (d_1(t), \beta_1(t) + ((i-1)/2)\cdot 2\pi/s_P)$ if $i$ uneven, $pos_t(r_i) = (d_2(t), \beta_2(t) + (i/2)\cdot 2\pi/s_P)$ if $i$ even.
		The robots form in the first round of the execution the pattern $P$ scaled to a diameter $\leq 1$. 
		From there on, they scale the pattern up.
		Therefore $\frac{d_1(t)}{d_2(t)} = \frac{D_1}{D_2}$ and $\beta_1(t) = \alpha_1, \beta_2(t) = \alpha_2$ for $t > 1$.
		It is clear, that the ''uneven'' ($i$ is uneven) robots on $pos_t(r_i) = (d_1(t), \alpha_1 + ((i-1)/2)\cdot 2\pi/s_P)$ can distinguish themself from the ''even'' robots on  $pos_t(r_i) = (d_2(t), \alpha_2 + (i/2)\cdot 2\pi/s_P)$ when $\frac{d_1(t)}{d_2(t)} = \frac{D_1}{D_2}$.
		This is enough to compute the global coordinate system.
		The robots move onto positions with the same angle and 
		$$d_1(i+1) = \min\left(d_1(i) + 1, d_1(i) \cdot \frac{d_2(i) + 1}{d_2(i)}, D_1\right)$$
		$$d_2(i+1) = \min\left(d_2(i) + 1, d_2(i) \cdot \frac{d_1(i) + 1}{d_1(i)}, D_2\right)$$
		This reached $d_1(t) = D_1$and $d_2(t) = D_2$ after $\leq \max(D_1, D_2) \leq |P|$ rounds.

\end{proof}
}
{\section{Auxiliary Results}%
\label{app:auxiliary}

\begin{lemma}%
\label{lem:3connectedsectors}
Consider a finite connected subset $S \subset \cR^2$ of symmetricity $s \coloneqq \sym(S) \in \N$ and size $\abs{S} \geq 3s$.
There exists a subset $C \subseteq S$ of size $\abs{C} = 3$ and with $\angle(C) < 2\pi / s$ such that $C$ is connected.
\end{lemma}
\begin{proof}
For $s = 1$ the statement is trivial.
So assume $s > 1$, such that $2\pi/s \leq \pi$.
Consider the points from $S$ in polar coordinates, where we consider angular coordinates on $\intoc{-\pi, \pi}$.
Since $S$ is connected and $\abs{S} \geq 3$, there are three points $p = (r, \phi), q_1 = (r_1, \phi_1'), q_2 = (r_2, \phi_2') \in S$ with $\dist(p, q_1) \leq 1$ and $\dist(p, q_2) \leq 1$.
Because $S$ has symmetricity $s$, without loss of generality we can assume $\phi \in \intco{0, 2\pi/s}$ (otherwise we find corresponding rotation-symmetric points for $p$, $q_1$, and $q_2$).
Using again the symmetricity, for each $i \in \set{1, 2}$ we find a $p_i = (r_i, \phi_i) \in S$ with $\phi_i \in \intco{0, 2\pi/s}$.

For $i \in \set{1, 2}$ and $k \in K \coloneqq \set{0, 1, \dots, s-1}$, define $p_{i, k} \coloneqq p_i + (0, k \cdot 2\pi/s) \eqqcolon \intoo{r_i, \phi_{i, k}}$ and set $S_i \coloneqq \set{p_{i, k} | k \in K} \subseteq S$.\footnote{
    Remember that we consider angular coordinates modulo the interval $\intco{-\pi, \pi}$, such that $\phi_{i, k} \in \intco{-\pi, \pi}$ for all $i \in \set{1, 2}$ and $k \in K \coloneqq \set{0, 1, \dots, s-1}$.
}
In other words, $S_i$ is the set of all points in $S$ that are rotation-symmetric to $p_i$.
In particular, $q_i \in S_i$.
For any $p_{i, k} \in S_i$, the distance formula for points in polar coordinates yields
\begin{dmath}%
\label{prf:3connectedsectors:polardistances}
\dist\intoo{p, p_{i, k}}^2
=
r^2 + r_i^2 - 2r \cdot r_i \cdot \cos\intoo*{\phi - \phi_i - k \cdot \frac{2\pi}{s}}
\end{dmath}.
By choice of $p_i$, we have $\abs{\phi - \phi_i} \in \intco{0, 2\pi/s}$.
The right-hand side of \cref{prf:3connectedsectors:polardistances} is minimized at the unique $k_i \in K$ for which $\abs{\phi - \phi_i - k_i \cdot 2\pi/s} \in \intco{0, \pi/s}$.
More exactly, we have
\begin{itemize}
\item $k_i = \phantomas[r]{s-1}{  0}$ if $                                \abs{\phi - \phi_i}  \leq \phantom{-2}   \pi/s$,
\item $k_i = \phantomas[r]{s-1}{  1}$ if $\phantomas{\abs{\phi - \phi_i}}{     \phi - \phi_i } \geq \phantom{- }  2\pi/s$, and
\item $k_i =                    s-1 $ if $\phantomas{\abs{\phi - \phi_i}}{     \phi - \phi_i } \leq              -2\pi/s$.
\end{itemize}
In any case, the choice of $k_i$ yields $\angle(p, p_i) = \abs{\phi - \phi_{i, k_i}} < \pi/s$ as well as (together with $q_i \in S_i \subseteq S$)
\begin{math}
\dist\intoo{p, p_{i, k_i}}
\leq
\dist\intoo{p, q_i}
\leq
1
\end{math}.

Now consider the three element set $C \coloneqq \set{p, p_{1, k_1}, p_{2, k_2}} \subseteq S$.
As shown above,
\begin{math}
\dist\intoo{p, p_{i, k_i}}
\leq
1
\end{math}
for both $i \in \set{1, 2}$, so $C$ is connected.
Moreover,
\begin{dmath}
\angle(C)
\leq
\angle(p, p_{1, k_1}) + \angle(p, p_{2, k_2})
<
2\pi/s
\end{dmath}.
Thus, we found a set $C$ with the required properties.
\end{proof}

\section{Measuring Precision}
\label{sec:measuring-precision-appendix}
We say a pattern 
$P = (p_1, p_2, \cdots)$ is formed in 
$c$-approximation if the final configuration 
$\cR = (r_1, r_2, \cdots)$ of robots can be translated/rotated such that 
$\sum_{i=1}^{|P|}dist(r_i, p_i) \leq c$.
We say robots move $\mu$-precise if they move on an arbitrary position inside a circle of radius $\mu$ around their computed target position. 
We still assume arbitrary precise measurement, but any inaccuracy in measurement can easily be translated to inaccuracy in movement.

It is clear that the movement of an $\epsilon$-granular drawing formation $F$ as defined in \Cref{def:df-move} still works with an $\mu < \frac{\epsilon}{2}$ precision such that robots can still identify the \emph{defining} robots of $F$ after a move.\footnote{A robot needs to search for a pair in distance $d, \epsilon - \mu < d < \epsilon+\mu$ and then find a third robot approximately forming a line with the pair.}
The anchor of $F$ will also have a $\mu$-precise movement in this case.
Because $F$ traverses a path of length $\Theta(|P|)$ the error will accumulate to $err = \Theta(|P| \cdot \mu)$.

In the proof \Cref{lem:drawing-full-pattern} we showed that a drawing formation is valid in certain conditions near to other robots.
This especially contains a condition, no drawing formation can be in distance $<\epsilon$ to $\partial C^\iindex$.
This, of course, includes the accumulative error in moving a drawing formation.
Therefore, $err$ must be smaller $\epsilon$.
$\mu = \cO(\epsilon \cdot |P|)$.

}
{\section{Pseudocode}
\label{sec:pseudocode}

The main algorithm is \textsc{PatternFormation} (\Cref{alg:main}).
It is described in the proof of \Cref{thm:main-result}.
Every robot executes this algorithm in each round.
For better readability, we pass the executing robot $r$ and the neighborhood in its viewing range to the algorithm.
All subroutines are derived from definitions, lemmas, and proofs of our work.
We reference the related definition/lemma next to the algorithm name.

\begin{algorithm}
	\caption{\textsc{PatternFormation}($P, r$, neighborhood) \hfill \Cref{thm:main-result}}
		\label{alg:main}
	\begin{algorithmic}
		\State $P^\einindex\gets$ symmetric component of $P$ \Comment \Cref{def:symmetric-component}
		\State $\bm{v}^\einindex \gets$ \textsc{ConstructDrawingPath}($P^\einindex$)
		\State $\phi \gets \min(\nicefrac{2\pi}{\sym(P)}, \pi/3)$
		\State $\Delta \gets 0.1$
		\State $\epsilon \gets \min(1/s_P, \mindist(P), 1/\sqrt{|P|}) \cdot c$ \Comment $c$ constant (see proof of \Cref{lem:existence-of-suitable-drawing-path})
		\State phase $\gets$ \textsc{ComputeCurrentAlgorithmPhase}($P, (\epsilon, \Delta, \phi), r$, neighborhood)
		\If{phase $=$ ``initial near gathering''}
			\State execute formation of initial drawing pattern (\Cref{def:initial-drawing-formation})
		\ElsIf{phase $=$ ``in drawing formation''}
			\State $F \gets$ \textsc{ComputeDrawingFormation}($(\epsilon, \Delta, \phi), r$, neighborhood, )
			\State \textsc{TraverseDrawingPath}($\bm{v}^\einindex, F, r$)
		\ElsIf{phase $=$ ``in intermediate drawing formation''}
			\State \textsc{ExecuteMoveOfLastThreeRobots}($\bm{v}^\einindex, r$, neighborhood)
		\ElsIf{phase $=$ ``dropped robot''}
			\State do nothing
		\EndIf
	\end{algorithmic}
\end{algorithm}

\begin{algorithm}
	\caption{\textsc{ConstructDrawingPath}($P^\iindex$) \hfill \Cref{lem:existence-of-suitable-drawing-path}}
		\label{alg:compute-drawing-path}
	\begin{algorithmic}
		\State $T^\einindex\gets$ drawing tree of $P^\einindex$ (\Cref{def:drawing-tree})
		\State $trav \gets$ traversal of $T^\einindex$
		\State $tail \gets$ \textsc{ConstructFittingTail($P^\einindex$)}
		\State \Return $traversal + tail$
	\end{algorithmic}
\end{algorithm}
 
\begin{algorithm}
	\caption{\textsc{ConstructFittingTail}($P^\iindex$)\hfill \Cref{lem:existence-of-suitable-drawing-path}}
		\label{alg:compute-drawing-path}
	
	The construction of the tail is defined in the beginning of the proof.
\end{algorithm}

\begin{algorithm}
	\caption{\textsc{ComputeCurrentAlgorithmPhase}($P, (\epsilon, \Delta, \phi), r$, neighborhood) \hfill \Cref{lem:phase-distinction}}
		\label{alg:algorithm-phase}
	
	We defined in \Cref{lem:phase-distinction} that we can distinguish four situations which we can label with ``initial near gathering'', ``in drawing formation'', ``in intermediate drawing formation'', and ``dropped robot''.
	In the proof, we defined how to distinguish these situations.
\end{algorithm}

\begin{algorithm}
	\caption{\textsc{ComputeDrawingFormation}($(\epsilon, \Delta, \phi), r$, neighborhood) \hfill \Cref{lem:valid-df}}
		\label{alg:compute-drawing-formation}
	\begin{algorithmic}
		\State $pairs \gets$ detect all pairs $(r_1, r_2)$ with $\dist(r_1, r_2) = \epsilon$ in neighborhood
		\For{all $(r_1, r_2) \in pairs$} 
			\State $r_3 \gets$ find colinear robot in distance $\leq \Delta$ to $r_1$ and $r_2$
			\If{$r_3$ exists}
				\State $a, \vec{d} \gets$ compute anchor and direction of the defining robots $r_1, r_2, r_3$ (\Cref{sec:df-implementation})
				\State $H \gets$ drawing hull with $a, \vec{d}, \Delta, \phi$ (\Cref{def:drawing-hull})
				\If{$r$ is in $H$}
					\State $F \gets$ $\epsilon$-granular drawing formation with hull $H$ (\Cref{def:epsilon-granular-df})
					\State \Return $F$
				\EndIf
			\EndIf
		\EndFor
	\end{algorithmic}
\end{algorithm}

\begin{algorithm}
	\caption{\textsc{TraverseDrawingPath}($\bm{v}^\einindex, F, r$) \hfill \Cref{lem:drawing-body}}
		\label{alg:traverse-drawing-path}
	\begin{algorithmic}
		\State $i \gets$ state of $F$ \Comment see paragraph ``Enumeration of States'' in the proof of \Cref{lem:drawing-body}
		\State let $\bm{v}^\einindex = (v_1, \cdots, v_k)$
		\If{$i < k$} \Comment anchor is not on the last node
			\State movement-vector $\gets$ $v_{i+1} - v_i$
			\If{$v_i \notin \tail(\bm{v}^\einindex)$}
				\State drop-coordinates $\gets \{p - v_i \textrm{ for } p \in \covered(v_i)\}$
			\Else 
				\State drop-coordinates $\gets \emptyset$
			\EndIf
			\State \textsc{MoveDrawingFormation}(move-vector, drop-coordinates, $i+1, F, r$)
		\EndIf
	\end{algorithmic}
\end{algorithm}

\begin{algorithm}
	\caption{\textsc{MoveDrawingFormation}(move-vector, drop-coordinates, $i, F, r$) \hfill \Cref{def:df-move}}
		\label{alg:move}
	
	We defined in \Cref{obs:moving} that a drawing formation $F$ can be moved.
	The notation above means that the anchor of $F$ is moved by relative vector ``movement-vector'' such that the drawing formation has state $i$ and robots are dropped on the coordinates ``drop-coordinates'' (also relative to the anchor).
\end{algorithm}

\begin{algorithm}
	\caption{\textsc{ExecuteMoveOfLastThreeRobots}($\bm{v}^\einindex, r$, neighborhood) \hfill \Cref{lem:drawing-tail}}
		\label{alg:move-last-three}
	
	We defined in \Cref{sec:drawing-tail-proof} how the last three positions of a symmetric component can be reached.
\end{algorithm}}
\clearpage
{\section{Example of Pattern Drawing}%
\label{sec:example-execution}

The following is an illustration of our protocol.
The target pattern (gray crosses) has symmetricity $1$ and the drawing formation $F$ has initial size $\ell = 7$ and state $i = 1$.
Black dots are robots (left: dropped robots; right: state robots), gray circles are vertices of the drawing path\footnote{
    The drawing path is for visualization only and \emph{not} the outcome of the construction in \cref{sec:dp-implementation}.
}.
On the left side an image of the current configuration with a symbolic representation of the $\epsilon$-granular drawing formation $F$ is depicted, while the right side zooms in on $F$, indicating the state $i$ of the drawing formation $F$ of size $\ell$ each time (see \cref{def:df-state-enum}).

\bigskip

\includegraphics[page=3]{figures/new/drawing-path.pdf}\hfill
\includegraphics[page=10]{figures/new/drawing-path.pdf}

\vspace*{12pt}
\includegraphics[page=5]{figures/new/drawing-path.pdf}\hfill
\includegraphics[page=11]{figures/new/drawing-path.pdf}

\vspace*{12pt}
\includegraphics[page=6]{figures/new/drawing-path.pdf}\hfill
\includegraphics[page=12]{figures/new/drawing-path.pdf}

\vspace*{12pt}
\includegraphics[page=7]{figures/new/drawing-path.pdf}\hfill
\includegraphics[page=13]{figures/new/drawing-path.pdf}

\vspace*{12pt}
\includegraphics[page=8]{figures/new/drawing-path.pdf}\hfill
\includegraphics[page=14]{figures/new/drawing-path.pdf}

\vspace*{12pt}
\includegraphics[page=9]{figures/new/drawing-path.pdf}
}

\end{document}